\documentclass[10pt,twocolumn,letterpaper]{article}

\usepackage{cvpr}              

\usepackage{graphicx}
\usepackage{amsmath}
\usepackage{amssymb}
\usepackage{booktabs}

\usepackage{epsfig}
\usepackage{tabularx}
\usepackage{makecell}
\usepackage{multirow}
\usepackage{amssymb}
\usepackage{amsmath}
\usepackage{gensymb}
\usepackage{paralist}
\usepackage{color, colortbl}
\usepackage{xspace}
\usepackage{url}
\usepackage{setspace}

\usepackage[pagebackref,breaklinks,colorlinks]{hyperref}

\usepackage[capitalize]{cleveref}
\crefname{section}{Sec.}{Secs.}
\Crefname{section}{Section}{Sections}
\Crefname{table}{Table}{Tables}
\crefname{table}{Tab.}{Tabs.}

\def\cvprPaperID{7121} 
\def\confName{CVPR}
\def\confYear{2022}

\usepackage{enumitem}
\usepackage{xcolor}
\usepackage{bm}
\usepackage{hyperref}
\usepackage{dsfont}

\newcommand{\tb}[1]{\textbf{#1}}

\definecolor{LightCyan}{rgb}{0.88,1,1}

\newcommand{\ourmodel}{\textsc{Marky-mT5}\xspace}
\newcommand{\spkfol}{SpkFol\xspace}
\newcommand{\spkfolstar}{SpkFol-RxR\xspace}

\newcommand{\pz}{\hphantom{0}}

\begin{document}

\title{Less is More: Generating Grounded Navigation Instructions from Landmarks}

\author{
{\normalsize \textbf{Su Wang \quad Ceslee Montgomery \quad Jordi Orbay \quad Vighnesh Birodkar \quad Aleksandra Faust}}\\
{\normalsize \textbf{Izzeddin Gur \quad Natasha Jaques \quad Austin Waters \quad Jason Baldridge \quad Peter Anderson} \vspace{5px}}\\
Google Research
}

\maketitle


\begin{abstract}
We study the automatic generation of navigation instructions from 360\degree{} images captured on indoor routes. Existing generators suffer from poor visual grounding, causing them to rely on language priors and hallucinate objects. Our \ourmodel system addresses this by focusing on visual landmarks; it comprises a first stage landmark detector and a second stage generator--a multimodal, multilingual, multitask encoder-decoder. To train it, we bootstrap grounded landmark annotations on top of the Room-across-Room (RxR) dataset. Using text parsers, weak supervision from RxR's pose traces, and a multilingual image-text encoder trained on 1.8b images, we identify 971k English, Hindi and Telugu landmark descriptions and ground them to specific regions in panoramas. On Room-to-Room, human wayfinders obtain success rates (SR) of 71\% following \ourmodel's instructions, just shy of their 75\% SR following human instructions---and well above SRs with other generators. Evaluations on RxR's longer, diverse paths obtain 61-64\% SRs on three languages. Generating such high-quality navigation instructions in novel environments is a step towards conversational navigation tools and could facilitate larger-scale training of instruction-following agents.
\end{abstract}


\section{Introduction}

Wayfinding--navigating to a destination--is an everyday task. 
We study the automatic generation of navigation instructions that effectively guide \textit{people}.
Template-based language generators that use cardinal directions and street names are commonly used in outdoor mapping applications, and some more flexible generation approaches rely on databases containing information about maps, roads and landmarks \cite{richter:klippel:2005, drager-koller-2012-generation, roth-frank-2010-computing}. In contrast, instructions for \textit{indoor} wayfinding require egocentric movement guidance and reference to the visual environment (e.g. notable objects). 

\begin{figure}[t]
\centering
\includegraphics[width=1.0\linewidth]{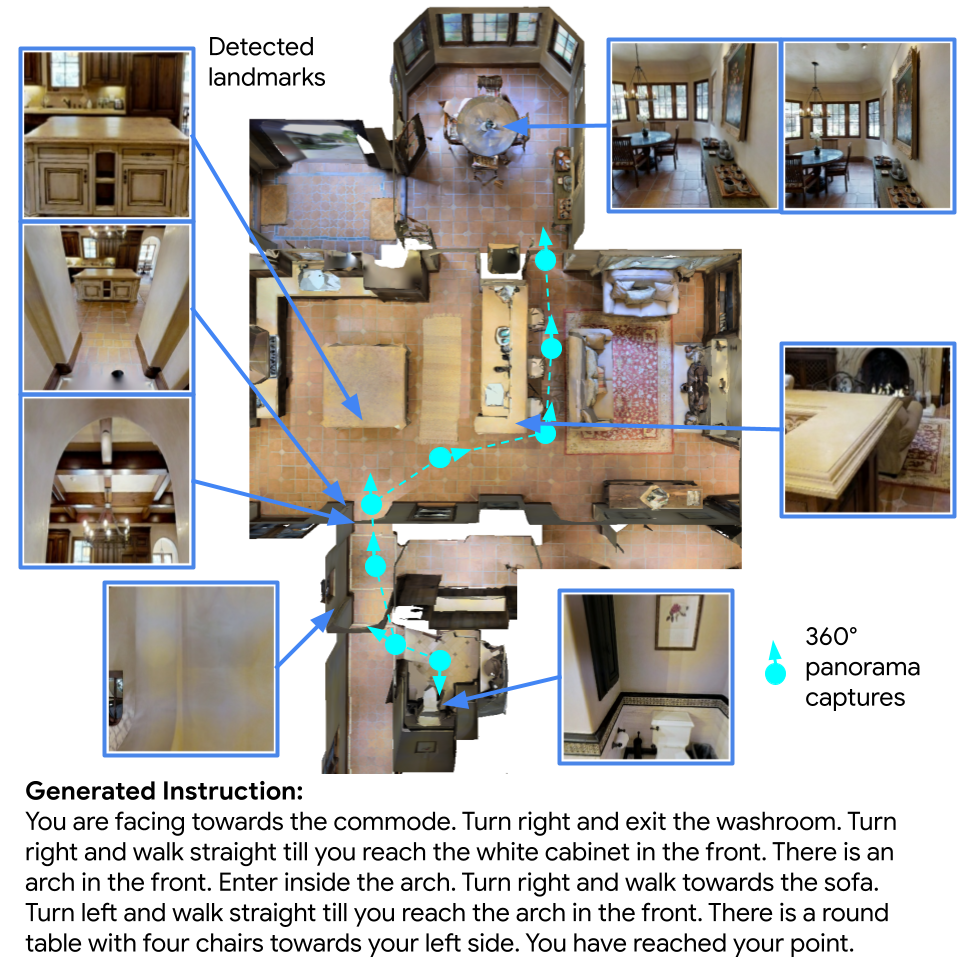}
\caption{We generate grounded navigation instructions from a sequence of 360\degree{} images captured along a route in a previously unseen building. Our two-stage approach first detects landmarks and then generates instructions conditioned on these landmarks.}
\label{fig:concept}
\vspace{-0.15in}
\end{figure}

Systems for generating indoor wayfinding instructions assume access to pre-existing floorplans and landmark databases~\cite{mast:wolter:2013}, but recent work attempts to generate novel instructions directly from visual inputs~\cite{fried2018speaker, backtranslate2019, Kurita21}. Progress toward this goal will enable navigation aids that are conversational rather than map-based---and it could provide a virtually unlimited supply of high-quality synthetic navigation instructions for training instruction-following robots. Describing navigation paths is also a key capability for human-robot communication, equipping robots to answer questions such as \textit{where did you go?} or \textit{where should I meet you?}.

We seek to generate accurate and fluent navigation instructions--in multiple languages--directly from visual representations and actions taken to traverse a path. Previous work assumed that the input to the instruction generator is a sequence of 360\degree{} panoramic (henceforth, \textit{pano}) images captured at intervals on a path, typically training on instructions from Room-to-Room (R2R)~\cite{Anderson:18} using Matterport3D environments~\cite{Matterport3D}. These models' instructions have proven valuable as additional training data for vision-and-language navigation (VLN) agents \cite{fried2018speaker}. However, \textit{people} struggle to follow them \cite{Zhao21}: human wayfinding success rates on R2R are 36\% for Speaker-Follower \cite{fried2018speaker} and 42\% for EnvDrop \cite{backtranslate2019} in unseen environments. The generated text is stylistically correct, but frequently references non-existent objects and confuses spatial terms such as left and right.

A challenge for visually-oriented instruction generators is dealing with irrelevant visual inputs. In many other image-to-text generation tasks (e.g., image captioning), much of the visual information in the input is reflected in the output text. This is not the case when generating navigation instructions. Human annotators look at less than 30\% of the environment~\cite{AlexKu:20}, and the instructions reference only a fraction of the objects that they look at. This makes learning a precise mapping between visual inputs and text outputs much harder. Perversely, access to more information can degrade performance~\cite{Haan:19}, as models happily learn spurious correlations that cause hallucinations during inference.

To solve this, we exploit the spatiotemporal grounding in the Room-across-Room (RxR) dataset~\cite{AlexKu:20}. Instead of \textit{writing} instructions, RxR annotators \textit{spoke} while traversing paths. Every RxR instruction thus comes with \textit{pose traces} that align the words spoken (and later transcribed) with what annotators were looking at. We use these pose traces and instructions to derive a new \textit{silver} annotated dataset\footnote{The term \textit{silver data} refers to high-quality annotations--not created by people--that are derived by combining models and constraints  \cite{rebholz-schuhmann-etal-2010-calbc,hahn-etal-2010-proposal,xia-etal-2021-stacked-amr}.} that contains bounding boxes over visual landmarks combined with their multilingual descriptions (English, Hindi and Telugu). Specifically, we bootstrap landmark annotations using text parsers to identify landmark phrases in instructions. We then use powerful image-text co-embedding models~\cite{mural:21} combined with weak supervision from pose traces to ground those landmarks in the environment. 

Our two-stage \ourmodel (land\textbf{mark} and \textbf{m}ultilingual \textbf{T5} \cite{xue-etal-2021-mt5}) system enhances instruction generation by improving how visual landmarks are selected and mentioned. Given a path-connected sequence of panoramic views, the first stage \textbf{landmark detector} infers a sequence of landmarks that a person might select for describing the path. E.g., in \cref{fig:concept} eight landmarks are selected, each represented by an image.
This sequence, plus interleaved descriptions of navigation actions, is passed to the second stage \textbf{instruction generator} -- a multimodal extension of the multilingual T5 (mT5) model \cite{xue-etal-2021-mt5} similar to VL-T5 \cite{vl-t5} -- to produce the instruction in \cref{fig:concept}.

In human wayfinding experiments on R2R paths, \ourmodel trained with silver landmarks (a subset of the visual inputs from the full environment) almost eliminates the gap between model-generated and human-written instructions -- achieving a 71\% success rate (SR) vs. 75\% for human instructions, 42\% for previous models, and 58\% for our model trained on full 360\degree{} panos. When it comes to selecting visual inputs for the generator, \textit{less is more}. On the more challenging RxR paths, human wayfinders obtain a 62\% SR using \ourmodel vs. 78\% for human instructions. We release our silver landmark data and over one million navigation instructions generated by \ourmodel, as data augmentation for training VLN agents.\footnote{\href{https://github.com/google-research-datasets/RxR/tree/main/marky-mT5}{github.com/google-research-datasets/RxR/tree/main/marky-mT5}}

\section{Related Work}


\noindent \textbf{Wayfinding with landmarks.} We wish to produce instructions that \textit{people} can follow, and are inspired by research on the importance of landmarks for human navigation \cite{Foo05,EdgarChan12,Epstein14,Yesiltepe21}. Landmarks are not just spatial features -- they encode a relation between the feature (e.g. object) itself, its nearby environment, and wayfinders' point of view \cite{caduff2008assessment}. Our landmark detector is trained on data bootstrapped from RxR's human-referenced landmarks. This allows our approach to exploit such features of landmark saliency, without designing for them explicitly (as in \cite{hile09, Fellner2017AutomaticGO}).

Hong \textit{et al.} \cite{NEURIPS2020_56dc0997} show that modeling relationships among the scene, its objects, and directional clues is effective for improving VLN wayfinding performance. This points to a potentially virtuous cycle between agents for wayfinding and those for guiding--or, better, using such landmark understanding for both capabilities within individual agents.

\noindent \textbf{Navigation instruction generation.}
Previous work on generating synthetic VLN instructions employs the Speaker-Follower framework \cite{Fried:18, backtranslate2019}:  A Speaker model learns from R2R annotations (English only) to produce instructions conditioned on sequences of path panoramas, while a Follower model learns to wayfind (i.e. construct paths) conditioned on human instructions and the same visual inputs. The Speaker's output can be used as augmented data for training the Follower, and the Speaker is used to rerank paths generated by the Follower during inference (a form of pragmatic reasoning).
These models indiscriminately use entire panos as visual context, whereas we select key visual landmarks from each pano for the generator to talk about. 
We build on the multitask, multilingual T5 model architecture \cite{2020t5, xue-etal-2021-mt5}, a unified text-to-text framework that enables transfer learning by mixing many NLP tasks simultaneously.  This additionally allows us to explore pretraining tasks, including multimodal tasks like image captioning, to improve generalization on unseen environments.  

Agarwal \textit{et al.} \cite{vls-2019} previously proposed a landmark-based generator, but relied on RL training rather than silver data to induce landmark groundings. Pashevich \textit{et al.} \cite{pashevich2021episodic} use synthetic instructions like \textit{goto bed pickup cellphone} as an additional source for training VLN agents for the ALFRED benchmark \cite{ALFRED20}. These are akin to the simple directional expressions we use in our multimodal encoder, but they are used both for data augmentation and as an additional decoding task. They seek to optimize VLN agent performance, whereas we seek to produce instructions that can be followed by people.

Kojima \textit{et al.} \cite{Kojima21} explore collaborative, human-agent instruction generation in the \textsc{CerealBar} game \cite{suhr-etal-2019-executing}. They define a human-in-the-loop instruction generation framework, where the generator is iteratively improved with signals collected while \textit{interacting} with people. The instructions cover both navigation and game tactics. 
 
Their multimodal$\rightarrow$text generator is a blackbox, whereas our approach includes an interpretable intermediate representation based on selected visual landmarks (compared to a formal abstraction, e.g. \cite{Daniele17}). This work complements ours---which only uses static human annotations---and suggests future interactive settings that could allow our instruction generators to adapt to human wayfinders.

\noindent \textbf{Multimodal generation.} Our use of landmarks and fine-grained connections between language and visual elements has parallels in image captioning. In particular, Pont-Tuset \textit{et al.} \cite{PontTuset_arxiv2019} show that controlled image captioning using mouse traces produces better image descriptions. 
Zhou \textit{et al.} \cite{Yuanen-Zhou21} use a pre-trained text-image matcher to learn keyword-bbox alignments, which are used to condition their caption generator (similar to our detection and use of visual landmarks). Huber et al. \cite{Huber18} provide a two-stage image-grounded dialogue model that extracts sentiment information from images to imbue emotion into generated texts. Specialized datasets, akin to our silver landmark data, are used to train an image feature extractor to improve scene understanding and handling of sentiment and facial features. 

Our generator itself falls within the growing collection of approaches that encode multimodal inputs and decode to text, including VL-T5 \cite{vl-t5}, MAnTiS \cite{sollami2021multimodal} and SimVLM \cite{wang2021simvlm}, often in a multitask setting. Our inputs are different in that we encode multiple, connected images that are interleaved with action descriptions. Our model outputs are evaluated with respect to downstream task performance using human evaluators, rather than automatic metrics and downstream tasks based on the learned representations.

\begin{figure}[t]
\centering
\includegraphics[width=0.9\linewidth]{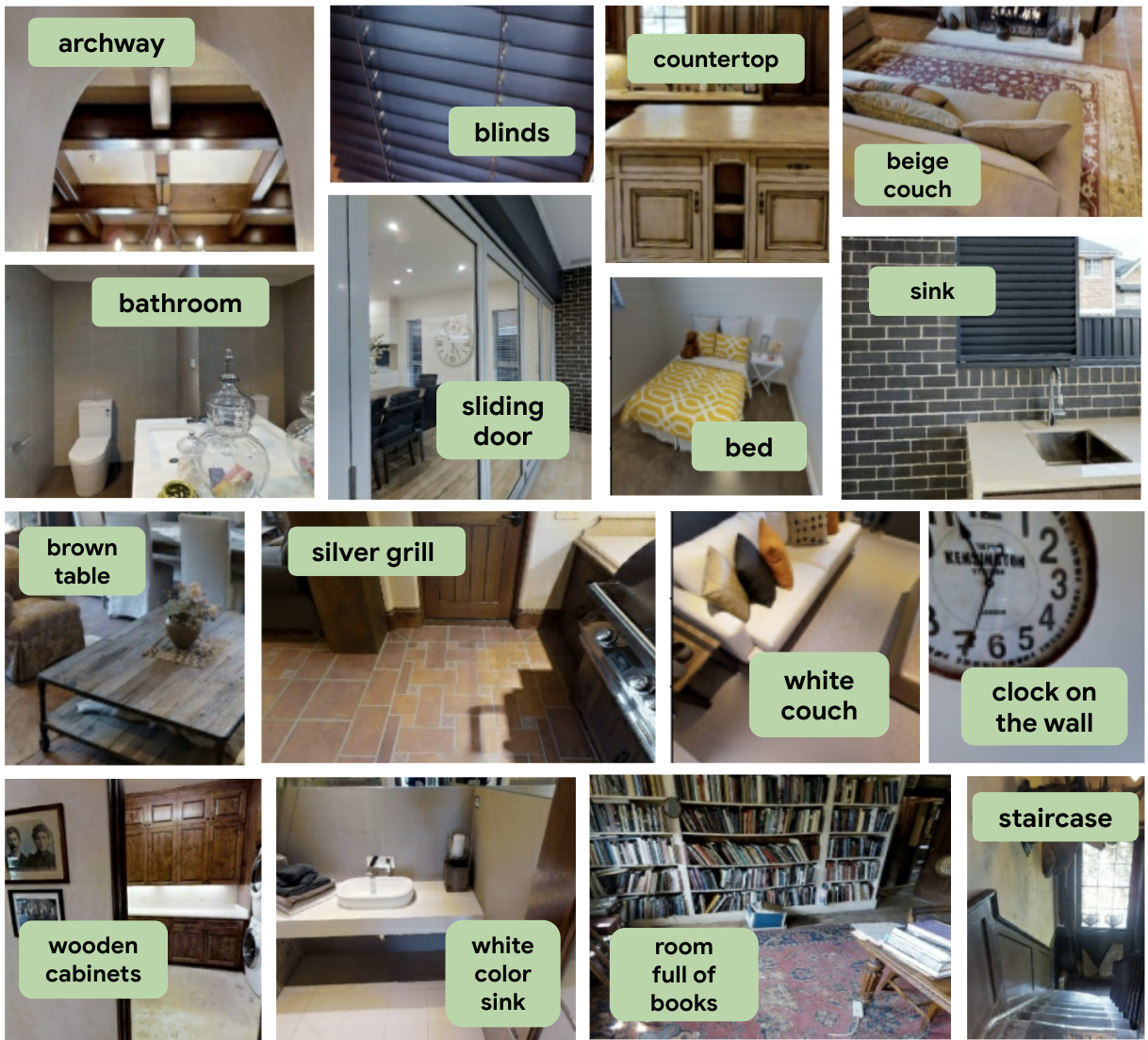}
\caption{Bootstrapped landmarks illustrated as image-text pairs.}
\label{fig:landmark-as-image-text-pair}
\vspace{-0.15in}
\end{figure}

\section{Bootstrapping a Landmark Dataset}
\label{sec:gold_landmarks}
\label{sec:silver_landmarks}

\ourmodel's first stage identifies visual landmarks as input to a second stage instruction generator. This requires training data of navigation instructions with landmark annotations. Here, a \emph{landmark annotation} is an instruction text span that mentions an object (e.g., \textit{white couch}), and the aligned perspective image containing that object (see \cref{fig:landmark-as-image-text-pair}). Unfortunately, navigation instruction datasets with annotated landmarks do not exist.\footnote{While Matterport3D contains 3D bounding box annotations for many objects (potential landmarks), these are not aligned to navigation instructions (which typically refer to a much broader set of landmarks than these).} Given how costly manual annotation would be at scale, we instead opt to bootstrap \textit{silver data}, a term used in natural language processing when high-quality annotations -- not created by people -- are derived by combining models and constraints \cite{rebholz-schuhmann-etal-2010-calbc,hahn-etal-2010-proposal,xia-etal-2021-stacked-amr}.

\noindent \textbf{RxR.} To automatically identify and annotate landmarks, we use \emph{Room-across-Room} (RxR) \cite{AlexKu:20}. RxR builds on the core task formulation defined by \emph{Room-to-Room} (R2R) \cite{Anderson:18}. Both are set in Matterport3D environments~\cite{Matterport3D}, but RxR is multilingual (English, Hindi, Telugu), its paths are longer and more varied, and it has roughly six times more instructions.  Importantly, while both R2R and RxR include navigation paths, RxR includes two key additional annotations:
\begin{compactitem}
    \item \emph{Pose traces}. A pose trace records an annotator's virtual pose as they move along a navigation path while recording spoken navigation instructions for others to follow. From a pose trace we can render a timestamped video of everything the annotator looked at.
    \item \emph{Text timestamps}. Every word in RxR's text instructions is timestamped, and thus aligned to the pose trace video. This provides a noisy, approximate grounding of words in the navigation instructions to video frames.
\end{compactitem}
\noindent
Using RxR, we construct landmark annotations in three steps: (1) \textit{extracting landmark phrases}; (2) \textit{grounding landmark phrases to pose trace frames}; and (3) \textit{refining selected frames to better align with the landmark object}. Text timestamps provide a key source of weak supervision for step 2, since an annotator would generally utter \emph{``white couch''} soon after seeing the couch. However, text timestamps alone are insufficient for grounding to pose trace frames as annotators may talk about a landmark without looking at it.

\noindent \textbf{Extracting landmark phrases.} The first step to create silver landmark data is extracting temporally-ordered lists of landmark phrases -- noun phrases describing objects in the environment -- from RxR's human instructions. 
For this, we use an mBERT-based~\cite{devlin2018bert} dependency parser trained on multilingual Wikipedia data~\cite{MultiWiki} to identify all entity mentions, including their parts-of-speech, which we consolidate with their non-clausal dependents into a single text span (see supplementary for further details). To improve results, in each language we manually curate a stop list of common landmark phrases that are discarded as too difficult to accurately localize (e.g., \textit{`the room'}, \textit{`your destination'}).  
The output is a text span sequence $\bm{t} = [t_1, t_2, \dots, t_m]$, e.g.
\begin{quote}
{\small \emph{You are standing in front of a \textbf{brown chair}. Take a left to enter the \textbf{bathroom}, you will see a \textbf{sink} in front of you. Now take a step to the right at stop at the \textbf{foot mat}, you will have reached your destination.} \\
$\rightarrow$ [\emph{\textbf{brown chair}, \textbf{bathroom}, \textbf{sink}, \textbf{foot mat}}]}.
\end{quote}
\noindent
In total, we extract 971k landmark phrases from the 102k instructions in the RxR Train, Val-Seen and Val-Unseen splits (9.55 landmarks per instruction on average). There are 84,237 unique phrases; the percentage of phrases that appear more than 3/5/10 times is 20.8\%/13.1\%/7.4\%. 

\noindent \textbf{Landmark grounding matrix.} For an instruction, to ground landmark phrases $\bm{t}{=}[t_1, t_2, \dots, t_m]$ to pose trace video frames $\bm{r}{=}[r_1, r_2, \dots, r_n]$,  we construct an $m{\times}n$ logit matrix $A$, where $A_{i,j}$ represents the compatibility of landmark phrase $t_i$ and frame $r_j$. Logits are computed by combining signals from MURAL-large~\cite{mural:21} -- a high-performing multilingual, multimodal dual encoder trained on a mixture of 1.8b noisy image-text pairs and 6b translation pairs -- and the RxR text timestamps, i.e.:
\begin{equation}\label{eq:logit}
    A_{i,j} = X(t_i) \cdot Y(r_j) - \lambda (T(t_i) - T(r_j))^2
\end{equation}
\noindent where $X$ computes a MURAL text embedding, $Y$ computes a MURAL image embedding, $T$ returns a timestamp in seconds, and $\lambda$ is a weighting factor set to 1.

\begin{figure}
\includegraphics[width=\columnwidth]{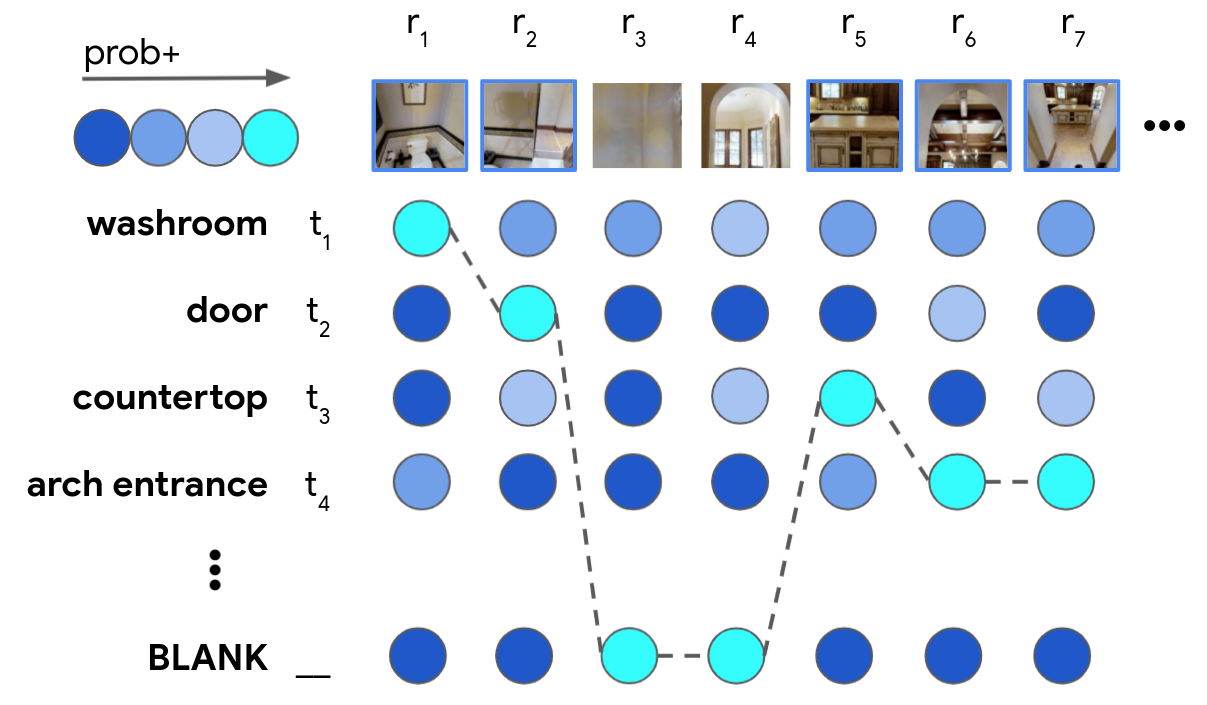}
\caption{Example CTC alignment between pose trace frames (top) and landmark phrases (left). Every landmark is aligned to a contiguous subsequence of frames. The $\mathtt{BLANK}$ label is used for frames that do not contain a landmark.}
\label{fig:ctc_alignment}
\vspace{-0.15in}
\end{figure}

To finetune MURAL on RxR, we add an additional trainable layer to each pretrained and frozen encoder and use \emph{Connectionist Temporal Classification} (CTC) loss~\cite{Graves:06}. Our motivation is similar to uses of CTC in speech recognition and optical character recognition~\cite{Xinjie-Feng18}, where the sequence of output labels is known, but the alignment to the inputs is not. We treat the frame sequence $\bm{r}{=}[r_1, r_2, \dots, r_n]$ as the input. The set of output labels $\{t_1, t_2, \dots, t_m, \mathtt{BLANK}\}$ includes the landmark phrases in the instruction plus a $\mathtt{BLANK}$ label (to accommodate frames without a landmark). The ordered landmark phrases $\bm{t}{=}[t_1, t_2, \dots, t_m]$ provide the target output sequence, i.e., we assume that the order in which annotators mention landmarks is the order in which they are observed. CTC's loss maximizes the probability that predicting a landmark phrase for each frame (or $\mathtt{BLANK}$) produces a valid output sequence:
\vspace{-0.1in}
\begin{equation}\label{eq:ctc}
    p(\bm{t}\mid \bm{r}) = \sum_{A^{(\bm{r}, \bm{t})}} \prod_{j=1}^{n} p(A_{i,j}\mid r_j)
\end{equation}
where $A^{(\bm{r}, \bm{t})}$ marginalizes over all valid alignments, and the right-hand side item computes the probability of single alignment. 
Intuitively, CTC performs training by encouraging each frame to be assigned to a landmark phrase such that the order of landmarks in the instruction is preserved. A valid CTC alignment is demonstrated in Figure \ref{fig:ctc_alignment}. 

\begin{figure}
\includegraphics[width=\columnwidth]{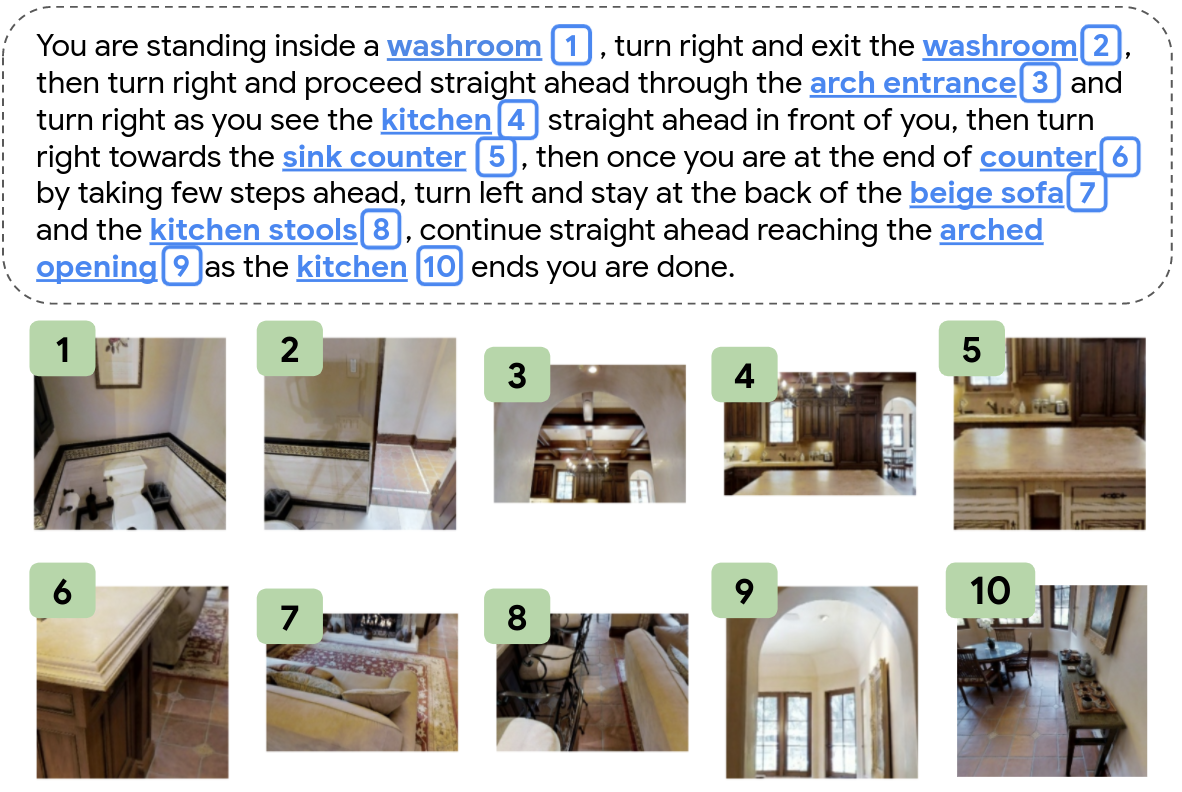}
\caption{A full bootstrapped landmark annotation for a navigation instruction from RxR Val-Unseen.
Nine of ten landmarks (1-9) are strongly aligned. Landmark 10 \emph{kitchen} is misaligned (despite the visual resemblance), perhaps due to a violation of our assumption that annotators mention landmarks in their observation order -- here, the mention of \emph{kitchen} harks back to an earlier observation.}
\label{fig:mp3d_alignment}
\vspace{-0.15in}
\end{figure}

\noindent \textbf{Decoding landmark alignments.}
After finetuning, we perform greedy CTC decoding on the logit matrix $A$ which aligns each landmark phrase with a sequence of one or more frames. For each landmark phrase $t_i$ we then find the best single alignment by selecting the frame in that sequence that assigns $t_i$ the highest probability. This generates image-text pairs where each text span is a landmark phrase from an instruction, and each image is a perspective projection from a 360\degree{} pano with known heading, pitch, and horizontal and vertical field of view (derived from the original pose trace).

\noindent \textbf{Evaluation.} As the text-image alignment task above is weakly-supervised (i.e., we know a landmark phrase sequence and a pose trace video are paired, but we do not know how landmark phrases are aligned with video frames), we also perform a small-scale human evaluation of alignment quality. We manually annotated 100 instructions from RxR Val-Unseen,
evaluating our approach in various conditions using alignment precision. We observe: 1) gains with adding timestamp-based bias (i.e. the squared term in \cref{eq:ctc}): 49.6 w/o bias vs. 56.5 w/ bias; 2) The bias alone (i.e. no input from MURAL) does not work: 23.4 precision.

\noindent \textbf{Landmark refinement.}
We observe that the pose trace frames selected in the previous steps typically contain the landmark phrase, but it is not always centered. This is because annotators do not necessarily need to look directly at a landmark to notice it. To further refine the landmark grounding, we generate a large number of alternative groundings (defined by heading, pitch and field of view) that overlap with the selected frame. The final silver landmark image for each landmark phrase is the grounding among these candidates with the highest MURAL alignment score with the landmark phrase. This results in very high quality landmark annotations, as shown in \cref{fig:mp3d_alignment} for a complete set of landmark annotations for one instruction.
\section{Landmark Detection}
\label{sec:landmark_detection}

Using navigation paths from RxR and silver landmark annotations (\cref{sec:silver_landmarks}), we train a \emph{landmark detector} that selects visual landmarks to be mentioned. The input to the model is a sequence of 360\degree{} panos $I_{1:T}$ captured at regular intervals along a route, along with the inbound and outbound direction. 
The model detects landmarks in 360\degree{} images as an object detection problem, using a CenterNet~\cite{XingyiZhou:06} object detector.
The detector is intended for identifying landmarks in unseen environments.
Although not all landmarks are objects in the traditional sense (e.g. \textit{open area}, \textit{laundry room}), previous work demonstrates that object detectors can learn to successfully detect and featurize amorphous image regions such as \textit{sky} and \textit{water}~\cite{Anderson2017up-down}.

\begin{figure}[t]
\centering
\includegraphics[trim=0 60 0 10,clip,width=1.0\linewidth]{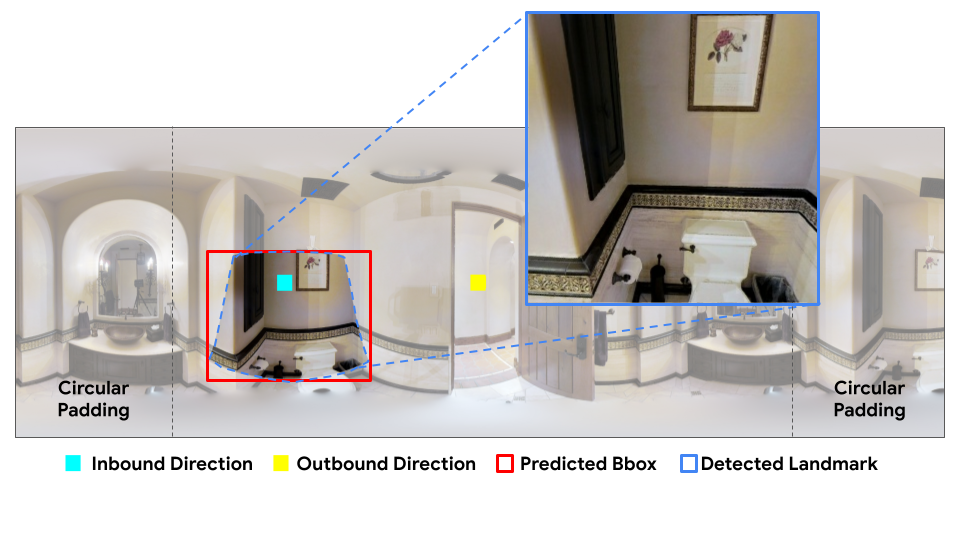}
\caption{Pano input format for the landmark detector. Both the inbound and outbound directions are drawn on the image to provide route context. The detector outputs bounding boxes in the equirectangular image, which are converted to perspective projections of the detected landmark(s).}
\label{fig:detector}
\vspace{-0.15in}
\end{figure}

\noindent \textbf{Input format.} Given a sequence of panoramic images $I_{1:T}$, each pano $I_t$ is input to CenterNet in equirectangular format. Though equirectangular images suffer from distortion, most landmarks are located close to the horizon, where equirectangular distortion is minimal.
Previous work on landmark salience for human navigation suggests that landmarks located closer to the next route segment are more important than landmarks further away (e.g. behind the observer)~\cite{10.1007/11556114_22,caduff2008assessment}. We therefore provide route context to the model, by rotating each input pano so that the direction to the next pano (the \textit{outbound direction}) is centered in the image. Since the direction faced on arrival from the previous pano (the \textit{inbound direction}) is also important, we draw colored pixel blocks on the input image indicating both to the model (see \cref{fig:detector}). To avoid edge discontinuities, we also add 90\degree{} of circular padding to each side of the pano. Images are prepared consistently for both training and inference.

\noindent \textbf{Training.} As outlined in \cref{sec:gold_landmarks}, each landmark image is a perspective projection from a 360\degree{} pano. To obtain training targets we map these projections to their minimum enclosing bounding box in the equirectangular pano image. We initialize the model with an Hourglass-104 backbone~\cite{cornernet,hourglass} pretrained on the COCO 2017 dataset~\cite{Lin2014}, and train for 20k iterations on silver landmarks from RxR training set using a batch size of 128, an equirectangular image size of 512{$\times$}256, and a single landmark object class.

\noindent \textbf{Inference.} During inference, we pool together the top 3 detections from each pano in the path, and return the top $T$ detections from this pool as the detected landmarks (where $T$ is the path length).\footnote{The 1:1 \#landmarks / path length ratio is selected based on BLEU and CIDEr scores: also experimented with 1.2:1, 1.5:1, 1.75:1 and 2:1 ratios.} These values (at most 3 landmarks from a single pano and one landmark per pano on average) closely correspond to averages in the training set. 

\section{Instruction Generation}
\label{sec:instruction_generation}

The instruction generator converts a representation of selected landmarks and the route connecting them into a textual navigation instruction. Our model is based on the mT5 text-to-text encoder-decoder transformer architecture~\cite{xue-etal-2021-mt5}, which is itself a multilingual variant of T5~\cite{2020t5} (hence, \ourmodel). In all experiments we use the mT5 base size model and the standard mT5 vocabulary which supports up to 101 languages using a SentencePiece~\cite{kudo2018sentencepiece} model trained on mC4 to encode text as WordPiece~\cite{wordpiece} tokens. 

\begin{figure}[t]
\centering
\includegraphics[width=1.0\linewidth]{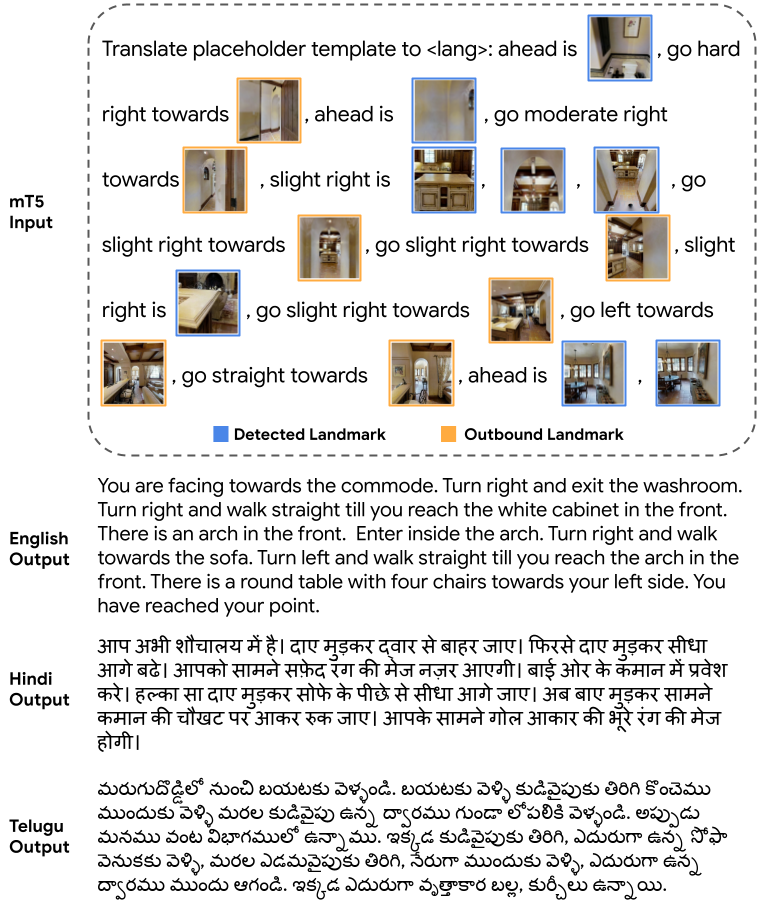}
\caption{Input template and multilingual outputs from the instruction generator. The input template encapsulates the visual landmarks to be mentioned along the route, along with their relative orientation and the navigation actions connecting them.}
\label{fig:generator}
\vspace{-0.15in}
\end{figure}

\noindent \textbf{Input format.} We formulate the input to the model by inserting visual representations of selected landmarks into a templatic English text sequence describing the orientation of each landmark and the actions required to traverse the route. Specifically, each visual landmark feature appears in order of its appearance on the route, preceded by its relative orientation to an observer, e.g. \textit{ahead is...}, \textit{slight right is...}. In between each landmark, the relative motion required to approach the next landmark is encoded with an action phrase, e.g. \textit{go hard left}, \textit{turn 180} (refer \cref{fig:generator}). There are 12 possible relative orientation phrases and 12 possible action phrases, with each one representing an arc of 30\degree{}. 
Using this approach, our instruction generator can be trained or evaluated using any set of landmark images that can be oriented along a 3D route. 
To identify the desired output language, the input is preceded by the appropriate prompt: \textit{`Translate placeholder template to English/Hindi/Telugu:'}. 

\noindent \textbf{Landmarks.}
To train the model we use the RxR dataset augmented with the silver landmarks described in \cref{sec:gold_landmarks}. 
Since instructions often mention the same landmark several times (e.g. from \cref{fig:mp3d_alignment} `You are standing in a \textit{washroom}, turn right and exit the \textit{washroom}...'), the same landmark image can appear multiple times in succession in the silver data. This leaks information about the surface realization of the instruction which will not be available from our landmark detector. We therefore de-duplicate landmarks by removing any landmark with heading, elevation and field of view within 5\degree{} of a previous landmark from the same pano.

In addition to the landmarks provided by silver data or the landmark detector (\textit{detected landmarks}), at each pano we include an additional \textit{outbound landmark}, defined as the view from the current pano in the direction of the next pano (bordered in orange in \cref{fig:generator}). Including outbound landmarks gives a small boost in automatic evaluations. All landmarks are represented using 640-dimension image embeddings from MURAL-large~\cite{mural:21}.

\begin{table*}[t]
\begin{center}
\scalebox{0.90}{
\small
\setlength\tabcolsep{3.5pt}
\begin{tabularx}{\linewidth}{llXcccccccccc}
&&&&&&&&& & \multicolumn{2}{c}{\textbf{Visual Search \%}} \\
 \cmidrule{11-12}
& & \textbf{Model}  &   \textbf{Landmarks} & \textbf{Training Data} & \textbf{WC} & \textbf{NE} $\downarrow$ & \textbf{SR} $\uparrow$ & \textbf{SPL} $\uparrow$  & \textbf{Quality} $\uparrow$   &  \textbf{Start} $\downarrow$  &  \textbf{Other} $\downarrow$ & \textbf{Time (s)} $\downarrow$ \\
\midrule
\multirow{6}{*}{\rotatebox[origin=c]{90}{\textbf{R2R (en)}}} &
1 & \spkfol~\cite{fried2018speaker} & Full Panos  & R2R & 24.6 & 6.0 & 42.0 & 35.8 & 4.1 & 39.8 & 23.6 & 54.2 \\
& 2 & EnvDrop~\cite{backtranslate2019}          & Full Panos & R2R & 24.5 & 6.0 & 41.7 & 35.3 & 4.0 & 40.7 & 23.5 & 54.0 \\
& 3 & \spkfolstar~\cite{fried2018speaker} & Full Panos  & RxR & 61.8 & 3.9 & 57.8 & 48.7 & 4.2 & 36.0 & 23.7 & 67.5 \\
& 4 & Marky-mT5      & Outbound & RxR & 57.5 & 3.6 & 64.9 & 54.1 & 4.2 & 36.2 & 23.8 & 72.5 \\
& 5 & Marky-mT5      & Predicted  &  RxR & 58.2 & 2.9 & 70.8 & 59.8 & 4.3 & 35.5 & 23.2 & 70.1 \\
& 6 & Human          & -  &  - & 25.6 & 2.8 & 74.9 & 66.4 & 4.5 & 37.8 & 23.0 & 52.2 \\
\end{tabularx}}
\end{center}
\vspace{-0.2in}
\caption{R2R Val-Unseen human wayfinding performance (\textbf{N} = 783 for each model). Combining the larger RxR dataset with landmark modeling and our bootstrapped landmark dataset, we almost eliminate the gap between model-generated and human-written instructions on paths of R2R-level difficulty -- achieving a 70.8\% success rate vs. 74.9\% for human instructions and 42\% for previous models.}
\label{tab:human-eval-r2r}
\end{table*}

\noindent \textbf{Rewrite task.}
To improve the performance of the instruction generator, we propose an auxiliary \textit{rewrite task}. In the rewrite task, we modify the input template during training by replacing the image feature for each silver landmark with the landmark phrase, e.g. `commode'. As before, the training target is a full navigation instruction containing all the landmark phrases in context. We hypothesize that this auxiliary task helps train the attention mechanisms in the model to correctly associate input landmark representations with their output realization in the text (noting that this is the only way we make direct use of the landmark phrase annotations from the silver dataset).

\noindent \textbf{Pretraining.}
We investigate pretraining the instruction generator on an equally weighted mixture of image-caption pairs from the CC3M~\cite{sharma-etal-2018-conceptual} and CC12M~\cite{Changpinyo_2021_CVPR} datasets. Since the captions are all in English, we translate them into Hindi and Telugu using an MT service.\footnote{\url{https://cloud.google.com/translate}}
We consider it important to introduce both text and visual inputs to our model during pretraining. 
We therefore formulate the pretraining task as machine translation of image captions using a pivot image~\cite{specia-etal-2016-shared}, e.g. (English $+$ image) $\rightarrow$ Hindi and so on.\ To encourage the model to make full use of the visual information only the first half of the input caption is provided.  

\noindent \textbf{Finetuning and inference.} We finetune mT5 on all three languages in the RxR train split using silver landmark inputs for 200k iterations with batch size 128. During inference we use argmax decoding and evaluate using both silver and predicted landmarks from the landmark detector.

\section{Experiments}
\label{sec:experiments}

\begin{table*}[t]
\begin{center}
\scalebox{0.90}{
\small
\setlength\tabcolsep{4pt}
\begin{tabularx}{\linewidth}{llXcccccccccc}
&&&&&&&&& & \multicolumn{2}{c}{\textbf{Visual Search \%}} \\
 \cmidrule{11-12}
& & \textbf{Model}  & \textbf{Landmarks} & \textbf{WC} & \textbf{NE} $\downarrow$ & \textbf{SR} $\uparrow$ & \textbf{SDTW} $\uparrow$ & \textbf{NDTW} $\uparrow$ & \textbf{Quality} $\uparrow$   &  \textbf{Start} $\downarrow$  &  \textbf{Other} $\downarrow$ & \textbf{Time (s)} $\downarrow$ \\
\midrule
\multirow{4}{*}{\rotatebox[origin=c]{90}{\textbf{RxR (all)}}} &
1 & Marky-mT5 & Outbound & 67.3 & 5.3 & 53.3 & 37.4 & 52.1 & 4.1 & 38.9 & 26.9 & 146.8 \\
& 2 & Marky-mT5 & Predicted & 71.1 & 4.4 & 61.5 & 43.9 & 56.2 & 4.2 & 37.6 & 26.5 & 154.4 \\
& 3 & Marky-mT5 & Silver & 72.2 & 4.3 & 63.3 & 45.4 & 57.0 & 4.3 & 37.8 & 26.6 & 149.8 \\
& 4 & Human          & - & 76.2 & 2.7 & 78.4 & 61.3 & 69.2 & 4.6 & 36.5 & 25.7 & 147.4 \\
\end{tabularx}}
\end{center}
\vspace{-0.2in}
\caption{RxR Val-Unseen human wayfinding performance (\textbf{N} = 4,551 for each model). On longer and more challenging RxR-style paths, human wayfinders obtain a 61.5\% success rate using our model vs. 78.4\% for human instructions. The higher 63.3\% success rate using silver landmarks suggests improved landmark selection is one way to narrow that gap.}
\label{tab:human-eval-rxr}
\vspace{-0.1in}
\end{table*}

\noindent \textbf{Baselines.} We train our multilingual \ourmodel model on RxR and evaluate using paths from both RxR\cite{AlexKu:20} and R2R\cite{mattersim}, comparing to the following prior work:
\begin{compactitem}
\item \textbf{\spkfol} and \textbf{EnvDrop}: Instruction generators from the Speaker-Follower~\cite{fried2018speaker} and Environmental Dropout~\cite{backtranslate2019} papers. 
Both models are LSTM\cite{Hochreiter1997} encoder-decoder models that take full panoramic images as input, with each pano represented by 36 image features representing different viewing directions.
\item \textbf{\spkfolstar}: We re-implement \spkfol and optimize hyperparameters for the longer instructions in the RxR dataset, training separate monolingual models for English, Hindi and Telugu using the mT5 vocabulary.
\end{compactitem}

\begin{table*}[t]
\begin{center}
\scalebox{0.90}{
\small
\setlength\tabcolsep{2.5pt}
\begin{tabularx}{\linewidth}{llXcccccccccccccc}
&&&&&&& \multicolumn{3}{c}{\textbf{Val-Seen}} & & \multicolumn{5}{c}{\textbf{Val-Unseen}} \\
\cmidrule{8-10} \cmidrule{12-16} 
&& \textbf{Model}  &  \textbf{Landmark} & \textbf{Rewrite} & \textbf{PT} & & \textbf{BLEU} & \textbf{CIDEr} & \textbf{SPICE (en)} & & \textbf{BLEU} & \textbf{CIDEr} & \textbf{SPICE (en)} & \textbf{SR} $\uparrow$ & \textbf{NDTW} $\uparrow$\\
\midrule
\multirow{7}{*}{\rotatebox[origin=c]{90}{\textbf{RxR}}} &
1 & \spkfolstar~\cite{fried2018speaker} & Full Panos & - & - &  & \pz5.9 & \pz8.9 & 13.9 &  & \pz5.7 & \pz8.4 & 13.0 &  29.6 & 41.6\\
& 2 & Marky-mT5 & Full Panos & - & - &  & \pz5.8 & \pz8.3 & 11.8 &  & \pz5.6 & \pz9.6 & 14.0 & 50.7 & 60.1\\
& 3 & Marky-mT5 & Outbound & - & - &  & \pz6.4 & 10.7 & 10.1 &  & \pz6.2 & 10.0 & 10.0 & 53.6 & 62.9\\
& 4 & Marky-mT5 & Silver & - & - &  & 11.6 & 25.2 & 16.6 &  & 11.1 & 23.3 & 15.7 & 55.9 & 64.1\\
& 5 & Marky-mT5 & Silver & \checkmark & - &  & 14.0 & 33.9 & \textbf{17.8} &  & 12.9 & 30.4 & \textbf{16.4} & 56.3 & \textbf{64.2}\\
& 6 & Marky-mT5 & Silver & \checkmark & \checkmark &  & \textbf{14.6} & \textbf{35.0} & 17.3 &  & \textbf{13.4} & \textbf{31.7} & 16.2 & 56.4 & \textbf{64.2}\\
& 7 & Marky-mT5 & Predicted & \checkmark & \checkmark &  & \pz6.2 & \pz8.4 & 13.7 &  & \pz5.8 & \pz7.5 & 13.5 & 55.7 & 63.3 \\
& 8 & Human & - & - & - &  & - & - & - &  & - & - & - & \textbf{56.5} & 62.9 \\
\end{tabularx}}
\end{center}
\vspace{-0.2in}
\caption{Automatic evaluations on RxR. Results are aggregated across English, Hindi and Telugu except for SPICE which is English-only. SR and NDTW score the wayfinding performance of a state-of-the-art VLN agent \cite{chen2021hamt}. PT refers to CC3M/12M PreTraining. 
}
\label{tab:auto-eval}
\vspace{-0.1in}
\end{table*}

\noindent
To evaluate modeling choices and opportunities, we report results for the following variations of our model:
\begin{compactitem}
\item \textbf{Outbound:} Training and evaluation with outbound landmarks, but no detected landmarks (refer \cref{sec:instruction_generation}). 
\item \textbf{Predicted:} Our full model, trained on silver landmark data and evaluated using predicted landmarks generated by our landmark detector.
\item \textbf{Silver:} As above, evaluated with silver landmarks. This allows us to assess a quasi-upperbound on performance on landmarks \textit{derived} from human annotations and does \textit{not} represent deployed system performance.
\item \textbf{Ablations:} Results without pretraining on CC3M and CC12M, and without the Rewrite auxiliary task.
\end{compactitem}

\noindent \textbf{Human Wayfinding.}
\label{subsec:human-wayfinding}
We consider human evaluations to be essential for this task. Using PanGEA~\cite{pangea}, an open-source annotation toolkit for panoramic graph environments, we immerse annotators in a simulated first-person environment backed by the Matterport3D dataset~\cite{Matterport3D} and ask them to follow the provided text navigation instructions. In total, we conduct 20k wayfinding evaluations involving 70 annotators. All annotators are based in India and are fluent in the language they were tasked with.

\noindent \textbf{Human Wayfinding Metrics.} We report the following standard metrics~\cite{evaluation2018,magalhaes2019effective} to evaluate the similarity between the route generated by our annotators and the intended route (and hence, the quality of the generated instructions): 
navigation error (\textbf{NE~$\downarrow$})
success rate (\textbf{SR~$\uparrow$}),
success weighted by inverse path length (\textbf{SPL~$\uparrow$}), 
normalized dynamic time warping (\textbf{NDTW~$\uparrow$}), and success weighted DTW (\textbf{SDTW~$\uparrow$}). Arrows indicate improving performance. 
Following \cite{Zhao21}, we also report instruction quality assessed by annotators on a 1--5 Likert scale (\textbf{Quality~$\uparrow$}), and the percentage of the available panoramic visual field that the annotator observes at each viewpoint (\textbf{Visual Search~$\downarrow$}), with higher values indicating greater effort spent looking for the correct route or landmarks.\footnote{\textbf{Start} and \textbf{Other} refer to first and other viewpoints, reported separately because wayfinders usually look around to orient themselves at the start.} \textbf{Time~$\downarrow$} represents the average time taken in seconds to complete the task, \textbf{WC} is word count and \textbf{N} is the number of instructions evaluated. 

\noindent \textbf{Automatic Metrics.}  For model development and ablations, we use automatic evaluation metrics. SPICE~\cite{spice2016} is the only metric found to correlate with human wayfinding performance~\cite{Zhao21}; however, SPICE needs an English dependency parser~\cite{klein-manning-2003-accurate}, so we also report BLEU~\cite{Papineni2002} and CIDEr~\cite{Vedantam2015}. Further, since SPICE was designed for evaluating individual sentences, not multi-sentence paragraphs, when calculating SPICE we separate both candidate and reference instructions into individual sentences (details in Supplementary). 
We compute CIDEr and SPICE with official evaluation code from COCO captions \cite{Chen2015}. For BLEU, we use the sacreBLEU implementation~\cite{post-2018-call}. CIDEr and BLEU scores are macro-averaged over languages.

\noindent \textbf{Results.} We present human evals on R2R's Val-Unseen paths in \cref{tab:human-eval-r2r} and RxR's Val-Unseen paths in \cref{tab:human-eval-rxr}; automated evaluations on RxR are in \cref{tab:auto-eval}. We align our results discussion to specific questions and themes.

\textit{Dataset choice:} Training on the larger RxR dataset instead of R2R noticeably improves human wayfinding performance in English, increasing success rate (42.0\% $\rightarrow$ 57.8\%) and SPL (35.8\% $\rightarrow$ 48.7\%) while lowering navigation error (6.0m $\rightarrow$ 3.9m) using the same model -- see \cref{tab:human-eval-r2r} row 3 vs. row 1. We note that the word count (WC) of models trained on RxR is considerably longer than R2R for the same paths; RxR annotations include more detailed descriptions and more state verification~\cite{AlexKu:20}.

\textit{Do landmarks matter?} Yes. On R2R, \ourmodel increases success rate (57.8\% $\rightarrow$ 70.8\%) and SPL (48.7\% $\rightarrow$ 59.8\%), and lowers navigation error (3.9m $\rightarrow$ 2.9m) compared to prior work without landmarks -- represented by \spkfolstar trained on the same dataset (\cref{tab:human-eval-r2r}, row 5 vs. 3). We attribute our gains primarily to landmark modeling: though \ourmodel uses a transformer architecture and stronger image features than \spkfolstar, these changes have little impact without landmarks (\cref{tab:auto-eval}, row 2 vs. 1).

\textit{How useful are the silver landmarks?} On RxR's challenging paths, training and evaluating with silver landmarks increases success rate (53.3\% $\rightarrow$ 63.3\%), increases NDTW (52.1\% $\rightarrow$ 57.0\%), and lowers navigation error (5.3m $\rightarrow$ 4.3m) compared to using only outbound landmarks (\cref{tab:human-eval-rxr}, row 3 vs. 1). This confirms our silver landmarks contain useful groundings and using them improves performance.

\textit{Can landmark detection be improved?} \ourmodel performs better using silver landmarks instead of predicted landmarks, e.g. success rate of 63.3\% vs. 61.5\% (\cref{tab:human-eval-rxr} row 3 vs. 2), suggesting gains are possible.

\textit{Comparison to human instructions:} Overall, using a combination of RxR data, silver landmarks and modeling improvements, we almost eliminate the gap between model-generated and human-written instructions on paths of R2R-level difficulty -- with a 71\% success rate vs. 75\% for human instructions and 42\% for previous models (\cref{tab:human-eval-r2r}). However, on the more challenging RxR-style paths, a gap remains -- human wayfinders obtain a 62\% success rate using \ourmodel vs. 78\% for human instructions (\cref{tab:human-eval-rxr}).
While humans can still distinguish \ourmodel instructions from human, a state-of-the-art VLN agent~\cite{chen2021hamt} cannot -- achieving near identical success rates (56.5\% vs. 55.7\%) and NDTW (62.9\% vs. 63.3\%) for human and model-generated instructions respectively (\cref{tab:auto-eval} row 8 vs. 7).


\textit{Rewrite auxiliary task and CC3M/12M pretraining:} Training with the Rewrite task (\cref{tab:auto-eval} row 5 vs. 4) has a large positive effect on Val-Seen and Val-Unseen BLEU (+2.4/+1.8), CIDEr (+8.7/+7,1) and SPICE (+1.2/+0.7). Pretraining (\cref{tab:auto-eval} row 6 vs. 5) improves BLEU (+0.6/+0.5) and CIDEr (+1.1/+1.3) scores, but not SPICE (-0.5/-0.2). 

\textit{Trustworthiness of automatic evaluations:} We note that replacing silver landmarks with predicted landmarks (\cref{tab:auto-eval} row 7 vs. 6) dramatically reduces BLEU (-8.2/-7.6), CIDEr (-26.6/-24.2) and SPICE (-3.6/-2.7) to the level of previous work. This is completely inconsistent with our human evaluations. We suggest \textit{only using automatic evaluations to compare similar models}, and attribute this reduction in textual similarity to differences in the choice of landmarks.

\noindent \textbf{Diverse generation.}
An appealing property of our two-stage approach is that diverse instructions can be generated by sampling landmark predictions.
Revisiting the \cref{fig:concept} example, sampling the next highest ranked landmark predictions produces the following--quite different--instruction:
\begin{quote}
{\small \emph{
You are standing in a bathroom facing a toilet. You are going to turn to your right and exit the bathroom. You will come into a kitchen area. There will be a marble counter top on your right and to your left a living room. You are going to make a left into the living room. In front of you will be an arched entrance way with a table and chairs. You are going to stop right in front that entrance way and you are done.
}}
\end{quote}

\section{Conclusion}


We propose a two-stage approach for generating navigation instructions directly from visual inputs. 
Supported by a new bootstrapped dataset of 971k grounded landmarks, \ourmodel almost eliminates the gap between model-generated and human-written instructions on R2R paths. Generating such high-quality navigation instructions in novel environments is a step towards conversational navigation tools and could facilitate larger-scale training of instruction-following agents. However, the strength of our approach--focusing on visual landmarks--is also it's limitation. \ourmodel is blind to environment context when generating, making it susceptible to pragmatic reasoning failures, e.g. generating \textit{`Leave the room'} in a room with multiple exits. Addressing this could lead to further gains.

\section*{Acknowledgements}


\noindent We thank Ming Zhao, Subhashini Venugopalan, and Alex Ku for early discussions and brainstorming; Yinfei Yang, Chao Jia and Aashi Jain for help with MURAL image features; Sebastian Goodman and Beer Changpinyo for assistance with the multimodal mT5 implementation; and Igor Karpov, Ming Zhao and the Google ML Data Operations team for support collecting human evaluations.


{\small
\bibliographystyle{ieee_fullname}
\bibliography{cvpr}
}

\clearpage

\section*{SUPPLEMENTARY MATERIAL}
\begin{figure*}[t]
\centering
\includegraphics[width=0.8\linewidth]{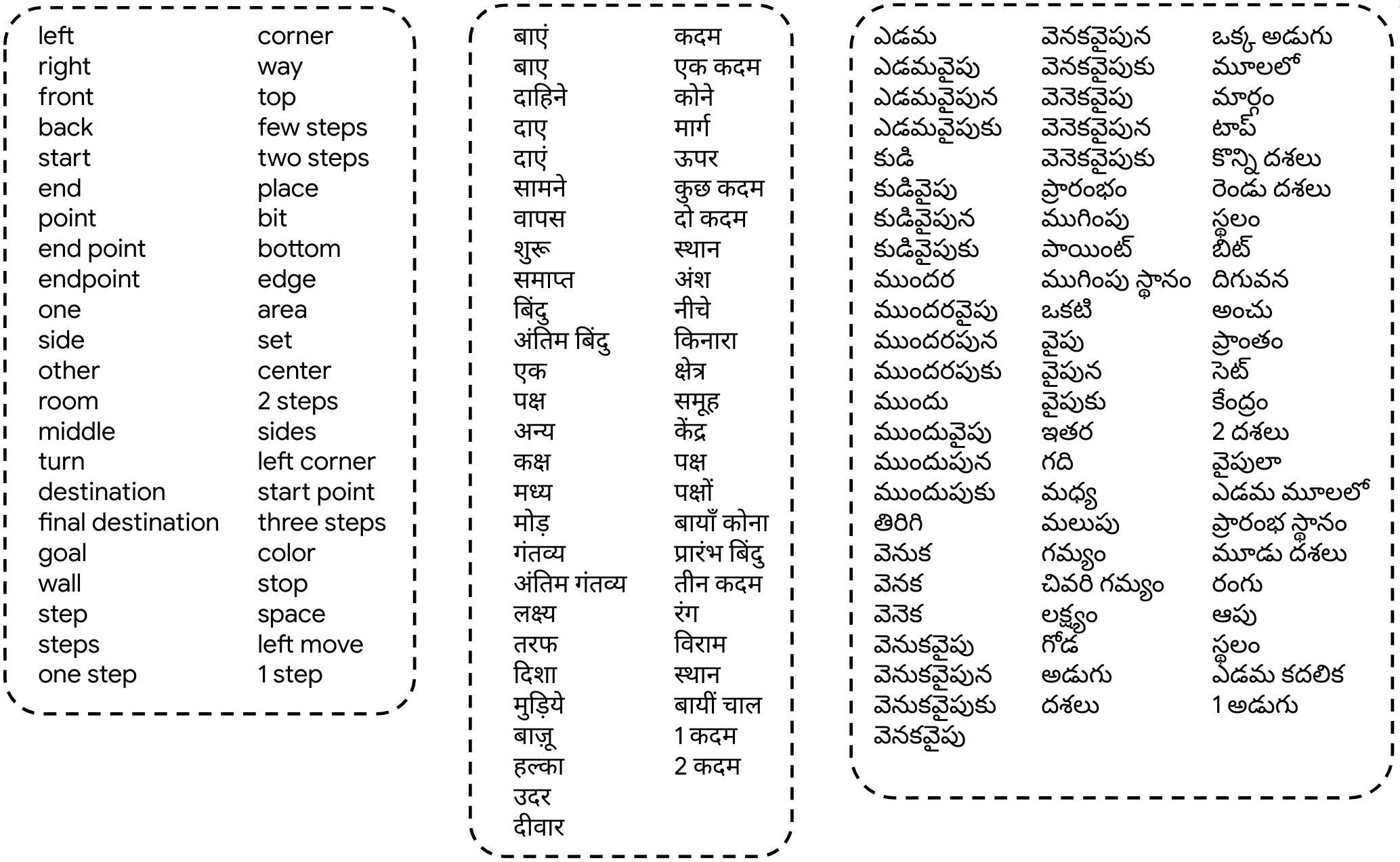}
\caption{Stoplists for excluding hard-to-localize noun phrases (NPs) when extracting landmark phrases. Left: English; Center: Hindi; Right: Telugu. Each stoplist was compiled by a native speaker examining samples of parsed landmark phrases.}
\label{fig:denylist}
\end{figure*}

We present supplementary material with section titles closely following the main paper.

\section{Bootstrapping a Landmark Dataset}
\label{supplementary}

\begin{figure*}
\centering
\includegraphics[width=0.8\textwidth]{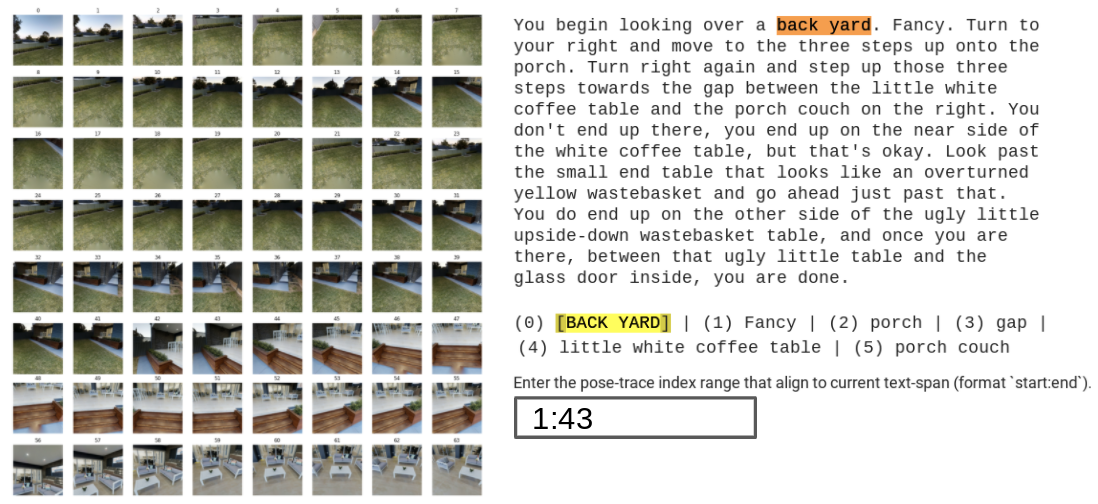}
\caption{Annotation interface used to collect ground-truth landmark groundings for a small-scale evaluation of the silver landmark dataset. For each automatically extracted landmark phrase (e.g., \textit{back yard}, right), the annotator inputs a range identifying the frames in the pose trace video (left) where that landmark can be seen.}
\label{fig:ls_annotation}
\end{figure*}

\paragraph{\textbf{Extracting landmark phrases.}}
\label{supp:parsing}
As outlined in Sec. 3, the first step to create silver landmark data is extracting temporally-ordered lists of landmark phrases from RxR's human instructions.
For this, we use a 3-layer distilled \cite{Tian-YongLong20} mBERT-based \cite{devlin2018bert} dependency parser pretrained on multilingual Wikipedia data \cite{MultiWiki} and outputting Universal Dependencies \cite{PengQi:18}. Specifically, the steps are:
\begin{enumerate}[itemsep=1mm, parsep=0pt]
    \item Extracting all entity mentions (along with their part-of-speech) from an instruction;
    \item Centering around the mentions as the head N/NP, absorbing all of their non-clausal dependents. E.g. for \emph{... at the large brown \tb{chair} that sits next to the window ...}, the mention is \emph{chair}, and the absorbed dependents are \emph{large} and \emph{brown} (but not the dependent clause \emph{that sits $\dots$});
    \item Consolidating the head N/NP and its absorbed dependents into a single text span and recording the indices of the first and last characters. In the example above, we produce \emph{large brown chair}.
\end{enumerate}
While the method identifies most desired landmark text spans, inevitable errors do occur, sometimes due to the imperfection of the human-written instructions, e.g.
\begin{itemize}[itemsep=1mm, parsep=0pt]
    \item Case 1. Non-object NPs. For \emph{``$\dots$ take two steps towards the door''}, for instance, \emph{steps} will be extracted as a candidate. Similarly, \emph{left} in \emph{``$\dots$ take a left''}.
    \item Case 2. Ill-formed/incomplete sentences. For example, in \cref{fig:ls_annotation}, the single-word sentence \emph{``Fancy.''} results in \emph{fancy} being parsed as a landmark text span.
\end{itemize}

To address Case 1, we manually compiled a \textit{stoplist} (\cref{fig:denylist}) of common non-object NPs that may appear in indoor environment (e.g. [\emph{side, step, one, endpoint}, $\dots$]) for each language (English, Hindi, Telugu). Case 2 is unavoidable with the tools currently available to us. Fortunately, they occur quite rarely. 

\begin{figure*}[!h]
\centering
\includegraphics[width=0.95\linewidth]{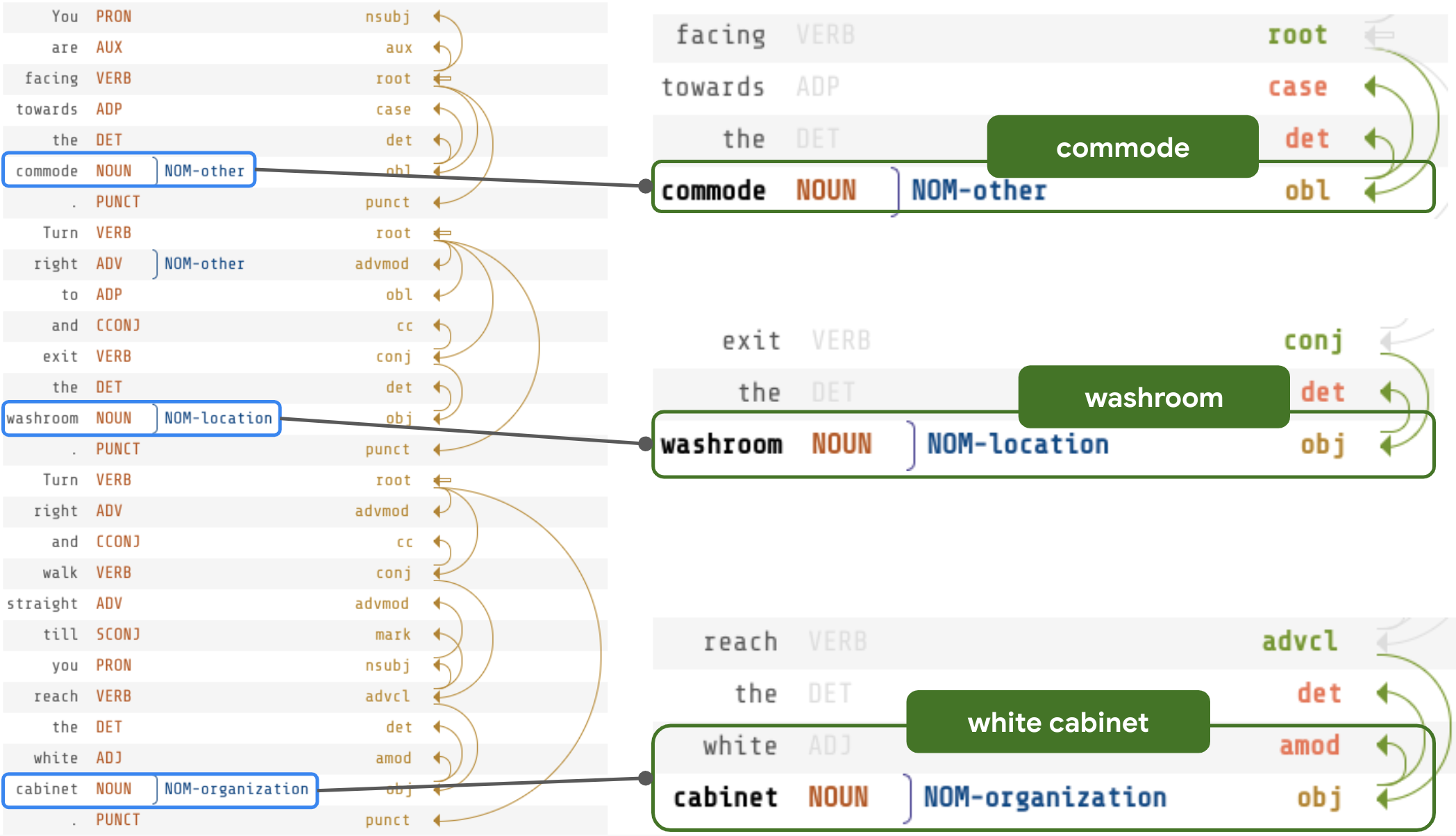}
\caption{Landmark phrase extraction for the instruction segment ``\emph{You are facing towards the commode. Turn right to and exit the washroom. Turn right and walk straight till you reach the white cabinet.}''. First the object-denoting heads are identified, then we filter through each one's dependency links to find full NP phrases for each head with constraints (e.g. only including adjectival, numeral, or adverbial dependencies, etc.).}
\label{fig:parsing-demo}
\end{figure*}

\cref{fig:parsing-demo} demonstrates landmark phrase extraction on an example instruction.

\paragraph{\textbf{Evaluation.}}
\label{supp:ls_annotation}
In Sec. 3 of the paper, we include results from a small-scale evaluation of 100 randomly-sampled English instructions from the silver landmark dataset. To provide ground truth landmark groundings, each automatically-extracted landmark phrase was manually aligned by the paper authors to a subsequence of frames from the corresponding pose trace video. The interface used for this manual-alignment is illustrated in \cref{fig:ls_annotation}. To compute the precision scores in the paper, whenever our automatic approach grounds a landmark phrase to a frame, it is a \textit{true positive} if the frame is in the human-selected subsequence, and a \textit{false positive} otherwise. Results are averaged over all landmark phrases in an instruction and then all instructions.

\begin{figure*}[t]
\centering
\includegraphics[width=1.0\linewidth]{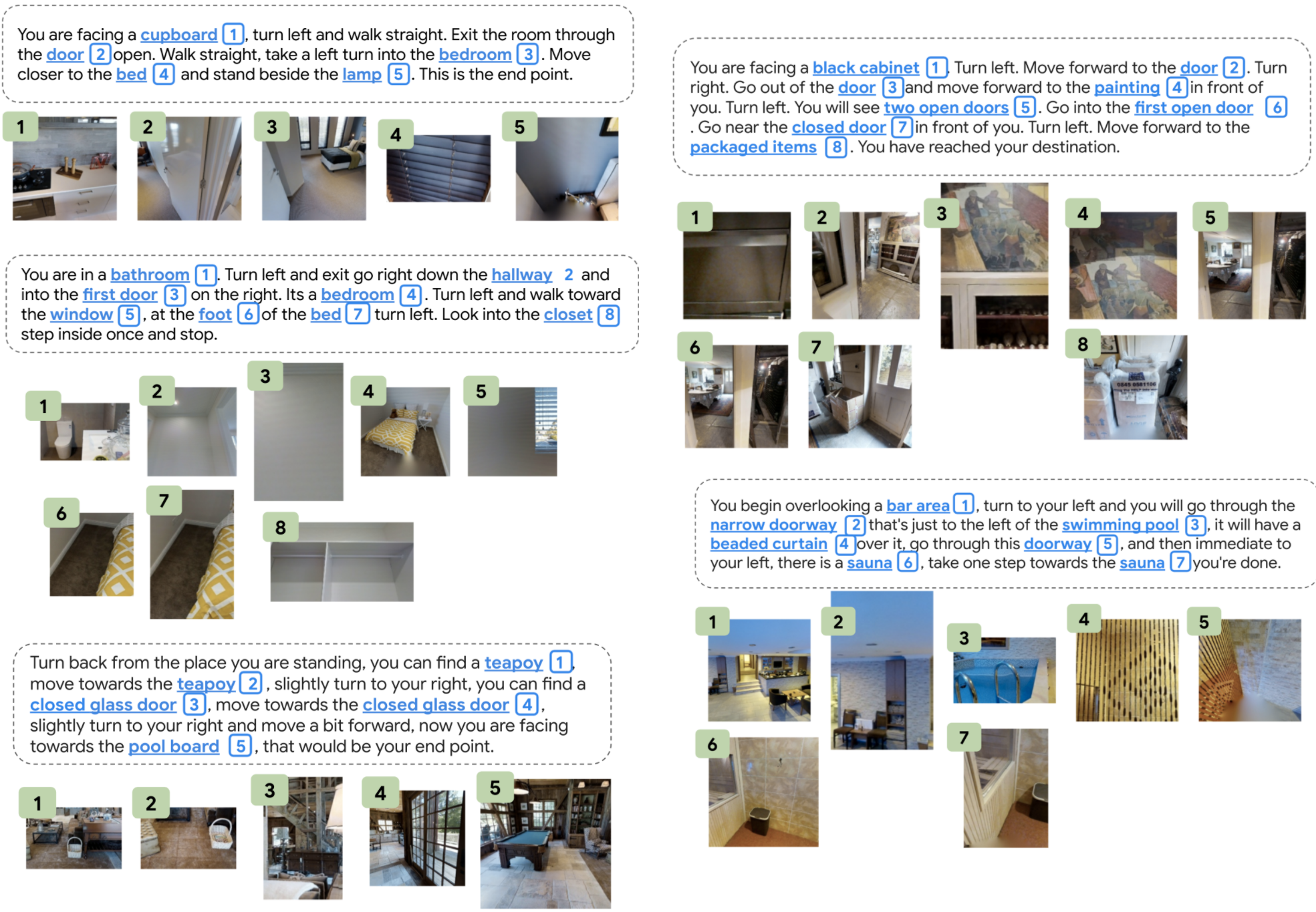}
\caption{Further examples of bootstrapped silver landmark annotations, illustrating both correct groundings (e.g., \textit{packaged items}, top right [8] and \textit{beaded curtain}, bottom right [4]) as well as failure cases (e.g., \textit{lamp}, top left [5]).}
\label{fig:mp3d-diverse}
\end{figure*}

\begin{figure*}[t]
\centering
\includegraphics[width=1.0\linewidth]{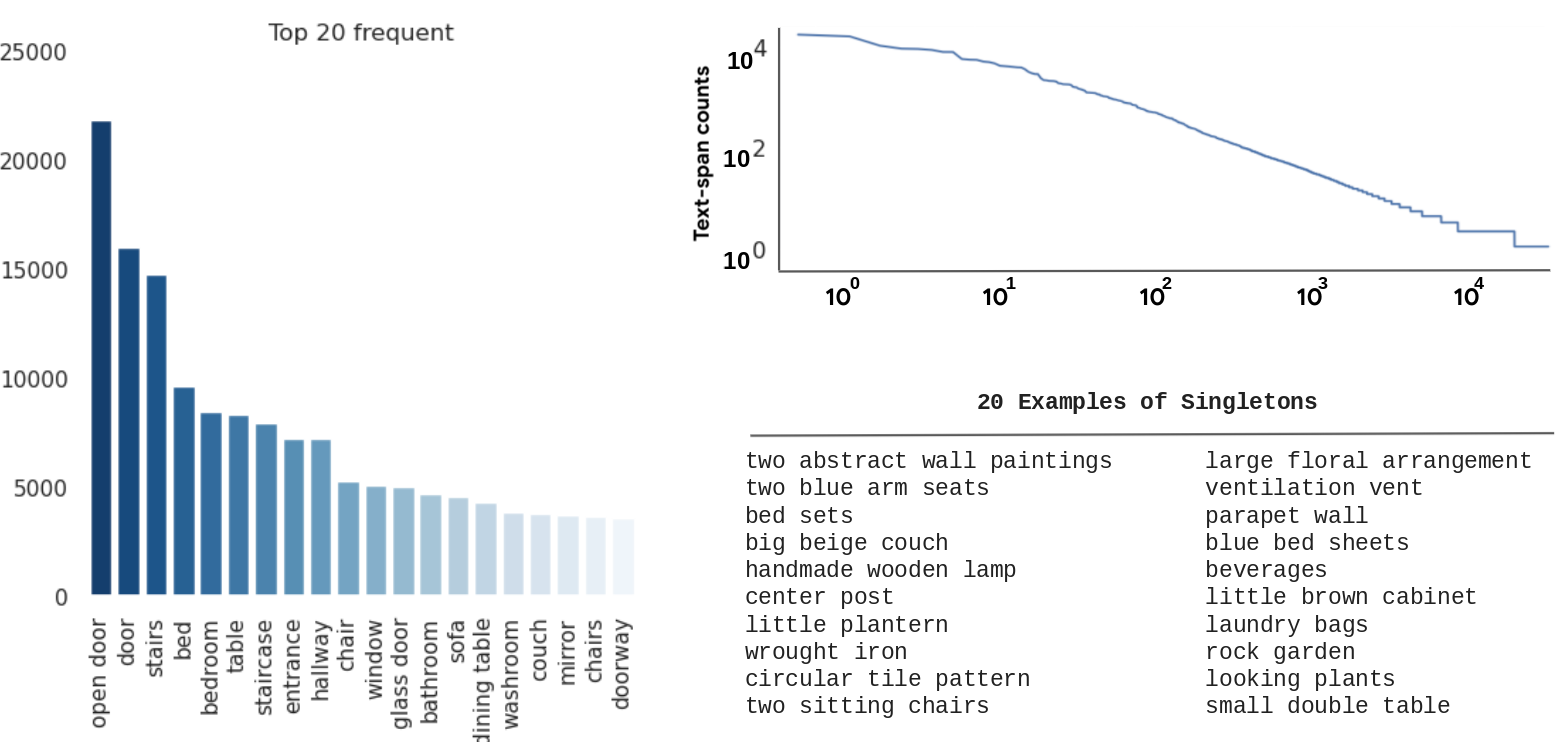}
\caption{\textbf{Left}: Frequency distribution of the top 20 landmark phrases. \textbf{Right top}: Long-tail distribution over all landmark phrases sorted in descending order (both x- and y-axes are labeled in log10-scale: landmark indices for x-axis, instance counts for y-axis). \textbf{Right bottom}: 20 samples of landmark phrases occurring only once. Common indoor/household items described generically (e.g. \emph{table} or \emph{door}) come on top of the rank over more specified ones (e.g. \emph{large brown dining table} or \emph{closed double door}). On the bottom end of frequency are specifically described uncommon objects, e.g. \emph{handmade wooden lamp}, \emph{parapet wall}, \emph{first two arched doorways}, etc.}
\label{fig:landmark-freq}
\end{figure*}

\begin{figure}[t]
\centering
\includegraphics[width=1.0\linewidth]{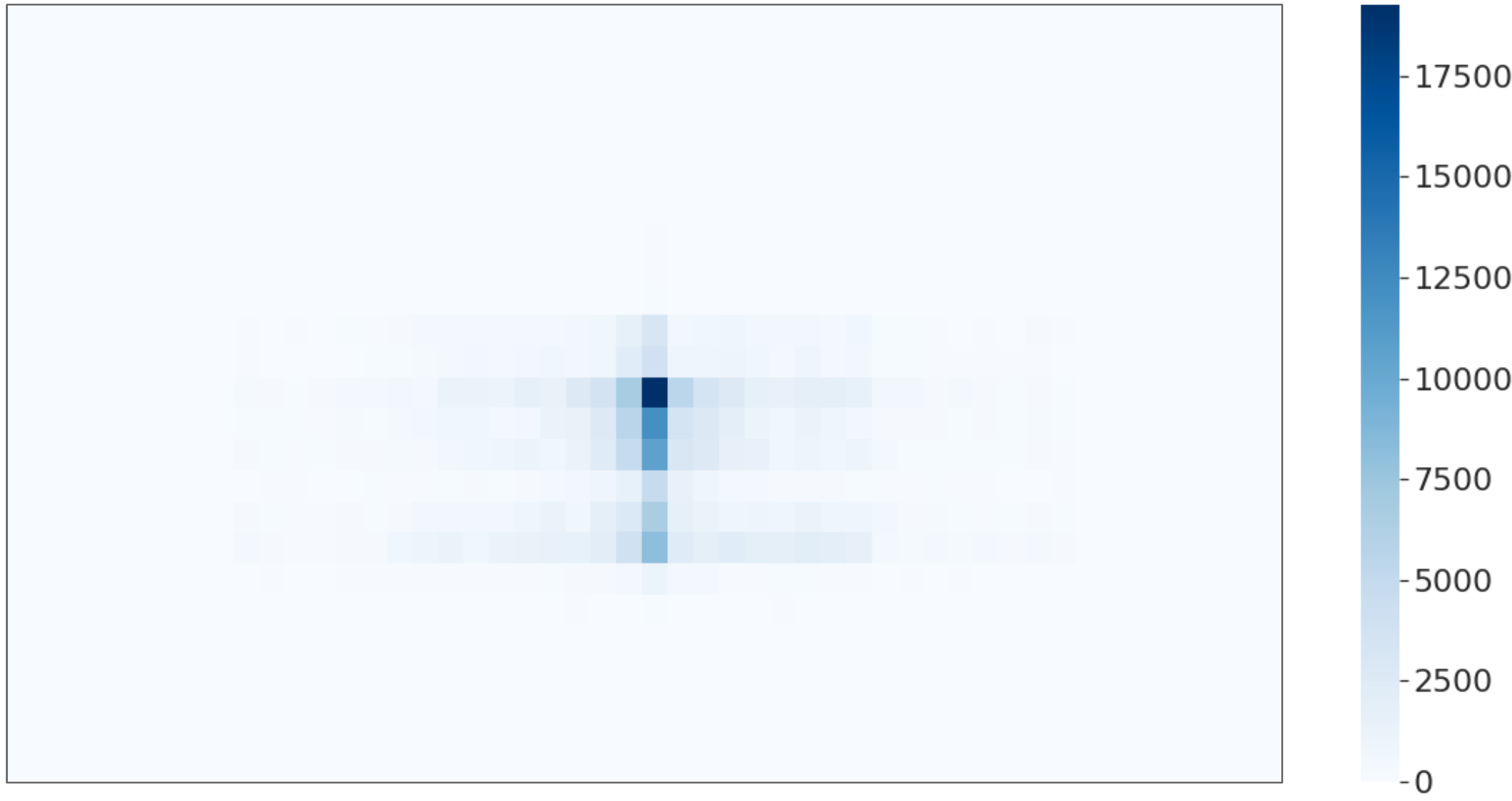}
\caption{Distribution of landmark centers in equirectangular image coordinates aligned to the outbound direction (i.e., the direction to the next pano on the route). As expected, landmarks are clustered around the outbound heading direction. With regards to pitch, most landmarks are found on the horizon or slightly below. The reason is fairly simple in the context of indoor environments: most landmarks are found on the floor or at table height, relatively few are found on walls and ceilings.}
\label{fig:landmark-wrt-outbound}
\end{figure}

\paragraph{Further analysis of the silver data.}
\cref{fig:mp3d-diverse} provides further examples of landmark annotations from our bootstrapped silver landmark dataset. 
\cref{fig:landmark-freq} presents the distribution of landmark phrases, including the 20 most frequent and least frequent landmark phrases. The distribution naturally has a long tail, although many unique landmark phrases are semantically similar (e.g. \emph{brown chair} vs. \emph{white chair}). We also empirically confirm the intuition that people tend to pick landmarks close to the \emph{outbound direction} (i.e. the direction towards the next route segment). \cref{fig:landmark-wrt-outbound} illustrates the distribution of landmark centers in equirectangular image coordinates aligned to the outbound direction of each pano. On the horizontal dimension (heading), the landmarks are clustered close to the outbound direction, whereas on the vertical dimension (pitch), landmarks are clustered on the horizon and slightly below.

\begin{table}[t]
    \centering
    \scalebox{.85}{
    \begin{tabular}{lcc}
        & BLEU & CIDEr   \\
        \toprule
        2-class & \textbf{5.3} & \textbf{7.7} \\
        10-class & 4.8 & 7.2 \\
        100-class & 4.4 & 6.4 \\
    \end{tabular}}
    \caption{Impact of training the landmark detector on clustered landmark classes. Automatic evaluation scores are reported for Marky-mT5 without pretraining, using the RxR Val-Unseen split.}
    \label{tab:detector-selection}
\end{table}

\section{Landmark Detection}
\label{supp:landmark-detection}

\paragraph{Training.} As noted in Sec. 4, the CenterNet landmark detector was trained using a single class to represent a landmark (i.e., 2 classes in total). In initial experiments, we also investigated training the detector using a richer set of landmark classes, by clustering 
landmark phrases (using $k$-means) into classes based on their MURAL-large~\cite{mural:21} text embeddings. However, this slightly reduced automatic evaluation scores for the full model (\cref{tab:detector-selection}), perhaps because detection confidence scores are uncalibrated across classes. 

\paragraph{Inference.} During inference with the landmark detector, we must determine how many landmarks to return. As described in Sec. 4, we used a 1:1 ratio of landmarks to path length, defined as the number of panos in the path. To determine this ratio, we experimented with the following values $\{1.0, 1.2, 1.5, 1.75, 2.0\}$ and computed automatic evaluations on the full model as reported in \cref{tab:num-lmk-path-len-ratio}.

\begin{table}[!ht]
    \centering
    \scalebox{.85}{
    \begin{tabular}{lcc}
        \#landmarks / path length ratio & BLEU & CIDEr   \\
        \toprule
        1.0 & \textbf{5.8} & \textbf{7.5} \\
        1.2 & 5.8 & 7.2 \\
        1.5 & 5.3 & 6.1 \\
        1.75 & 4.9 & 5.0 \\
        2.0 & 4.4 & 3.7 \\
    \end{tabular}}
    \caption{Impact of varying the \#landmarks / path length ratio. Automatic evaluation scores are reported for Marky-mT5 with Rewrite auxiliary training and CC3M/12M pretraining.}
    \label{tab:num-lmk-path-len-ratio}
\end{table}

\section{Instruction Generation}

\paragraph{Pretraining and Finetuning.} During pretraining, models are trained with Cross Entropy Loss and optimized with Adafactor \cite{shazeer2018adafactor} with a learning rate of 1.  Batch size is 128.  Pretrained models were trained for 1.45M steps.  Each pretrained model (CC3M, CC12M, CC3M+CC12M) was fine-tuned and the final pretrained version of the downstream model was selected based on SPICE.

\begin{table*}[t]
\begin{center}
\small
\setlength\tabcolsep{4pt}
\begin{tabularx}{\linewidth}{llXcccccccccc}
&&&&&&&&& & \multicolumn{2}{c}{\textbf{Visual Search \%}} \\
 \cmidrule{11-12}
& & \textbf{Model}  & \textbf{Landmarks} & \textbf{WC} & \textbf{NE} $\downarrow$ & \textbf{SR} $\uparrow$ & \textbf{SDTW} $\uparrow$ & \textbf{NDTW} $\uparrow$ & \textbf{Quality} $\uparrow$   &  \textbf{Start} $\downarrow$  &  \textbf{Other} $\downarrow$ & \textbf{Time (s)} $\downarrow$ \\
\midrule
\multirow{4}{*}{\rotatebox[origin=c]{90}{\textbf{RxR (en)}}} &
1 & Marky-mT5 & Outbound & 75.0 & 5.2 & 52.8 & 40.0 & 57.0 & 4.2 & 36.7 & 25.4 & 101.8 \\
& 2 & Marky-mT5 & Predicted & 81.6 & 4.2 & 61.3 & 46.5 & 61.2 & 4.3 & 36.1 & 24.6 & 107.6 \\
& 3 & Marky-mT5 & Silver  & 91.2 & 4.0 & 65.0 & 49.0 & 60.8 & 4.3 & 36.3 & 25.2 & 118.4 \\
& 4 & Human      &  -  & 98.6 & 2.7 & 77.5 & 62.2 & 71.0 & 4.6 & 35.3 & 24.3 & 113.5 \\
\midrule
\multirow{4}{*}{\rotatebox[origin=c]{90}{\textbf{RxR (hi)}}} &
1 & Marky-mT5 & Outbound & 82.1 & 5.6 & 50.8 & 32.6 & 46.4 & 4.2 & 42.5 & 27.4 & 194.8 \\
& 2 & Marky-mT5 & Predicted & 78.5 & 4.7 & 59.4 & 40.9 & 52.5 & 4.2 & 38.4 & 27.6 & 192.4 \\
& 3 & Marky-mT5 & Silver  & 72.4 & 4.6 & 60.7 & 42.3 & 54.4 & 4.3 & 39.1 & 27.1 & 176.0 \\
& 4 & Human      &  -  & 75.0 & 2.9 & 77.1 & 59.4 & 67.8 & 4.6 & 37.0 & 26.3 & 171.5 \\
\midrule
\multirow{4}{*}{\rotatebox[origin=c]{90}{\textbf{RxR (te)}}} &
1 & Marky-mT5 & Outbound & 43.2 & 5.2 & 56.3 & 39.6 & 52.9 & 4.1 & 37.5 & 27.9 & 143.8 \\
& 2 & Marky-mT5 & Predicted & 52.1 & 4.3 & 63.9 & 44.6 & 55.1 & 4.1 & 38.3 & 27.2 & 162.5 \\
& 3 & Marky-mT5 & Silver & 51.9 & 4.4 & 64.2 & 45.0 & 55.9 & 4.1 & 38.1 & 27.5 & 154.5 \\
& 4 & Human          & - & 53.7 & 2.6 & 80.9 & 62.6 & 68.9 & 4.4 & 37.1 & 26.5 & 157.4 \\
\bottomrule
\end{tabularx}
\end{center}
\vspace{-0.2in}
\caption{RxR Val-Unseen human wayfinding performance (\textbf{N} = 1,517 for each model), reported separately by language (English, Hindi and Telugu).}
\label{tab:human-eval-rxr-by-lang}
\end{table*}

\section{Experiments}

\paragraph{Human Wayfinding}

\begin{figure}[!t]
\centering
\includegraphics[width=1.0\linewidth]{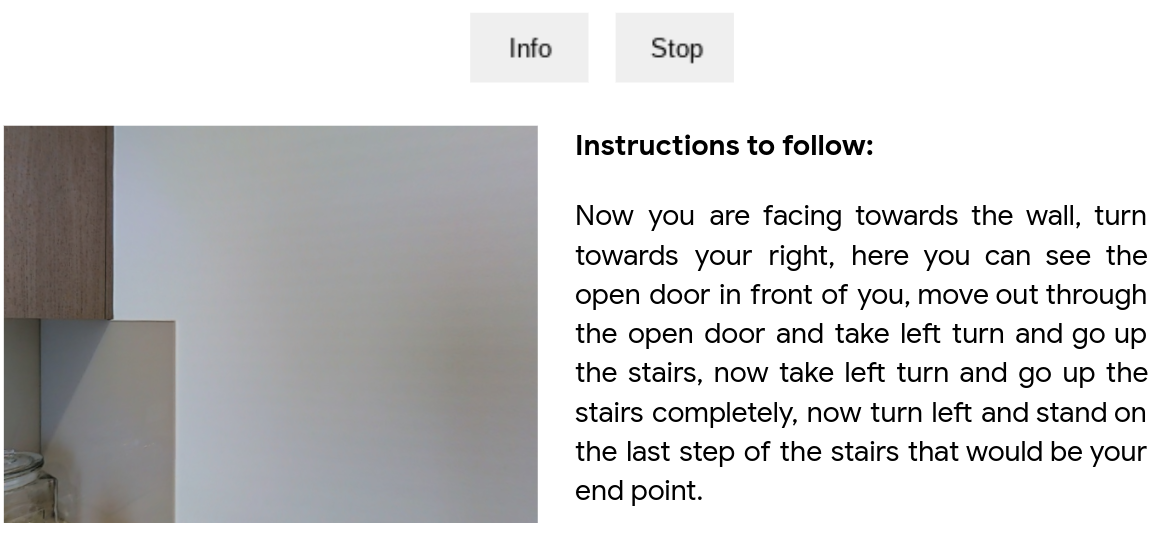}
\includegraphics[width=1.0\linewidth]{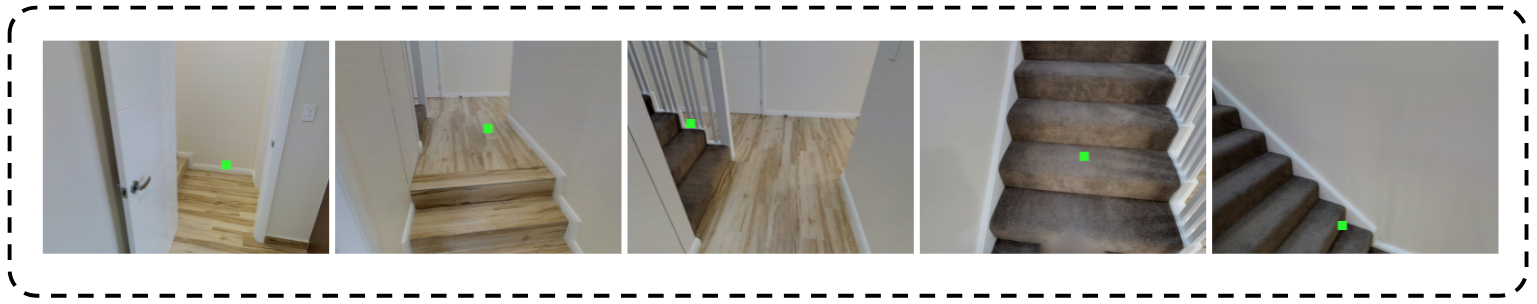}
\caption{Top: the PanGEA interface used in human wayfinding evaluations; Bottom: illustration of a series of first person snapshots taken along a navigation path.}
\label{fig:pangea-ui}
\end{figure}

In the PanGEA interface, each annotator is shown a virtual environment in a window on the left, paired with the textual navigation instruction being evaluated on the right (refer top pane in \cref{fig:pangea-ui}). Hovering their mouse on the window, the annotator will see a green-square indicator showing them the next locations available for them to move to. After double clicking on the green-square, they are taken to the next location --- illustrated in \cref{fig:pangea-ui} bottom pane, which presents a chain of first person snapshots taken while moving. Upon arriving at the location the annotator believes is the destination,  they may hit the \texttt{STOP} button to finish the task. 

Afterwards, the annotator responds to a multi-choice question to provide a subjective rating of instruction quality, classifying the instruction as containing:
\begin{compactitem}
    \item \emph{No mistakes, very very easy to follow};
    \item \emph{Few mistakes, easy to follow};
    \item \emph{Some mistakes, but still not hard to follow};
    \item \emph{Many mistakes, hard to follow};
    \item \emph{Way too many mistakes to follow}.
\end{compactitem}
From the top to the bottom, the answer determines the \textit{Quality} metric (a Likert score from 5 to 1). All annotators were fluent in the languages in the instructions given to them for wayfinding. The annotators were paid hourly wages that are competitive for their locale, and they have standard rights as contractors.

\paragraph{Results} 
In Tab. 2 of the main paper we report human wayfinding results on paths from the RxR Val-Unseen split, aggregated overall all languages. In \cref{tab:human-eval-rxr-by-lang} we report these results separately for each language (English, Hindi and Telugu). Results -- and most importantly, the patterns holding between the different settings -- are consistent across all languages.

Finally note that, for both R2R and RxR evaluation (Val-Unseen), each path is only evaluated once per language. Therefore, for R2R we have 783 items/paths, for RxR, we have 1,517 paths, and with 3 languages, 4,551 items.

\paragraph{Error analysis on Marky-mT5 instructions.}
Our human wayfinding results are representative of a step-change in instruction quality compared to previous models. To better understand the types of errors that still remain, we perform a manual error analysis on 110 randomly-sampled English instructions generated for RxR Val-Unseen paths by the full Marky-mT5 system using predicted landmarks.

To perform the analysis we added an option in the PanGEA wayfinding interface to toggle the visibility of the ground-truth path, so that we could better assess how well that path was described by the generated instruction. 
We classify instruction errors and weaknesses into the following six categories:
\begin{compactitem}
\item \textbf{Landmark Errors}
    \begin{compactitem}
    \item \emph{Full Hallucination}. The instruction describes a landmark that does not exist in the environment;
    \item \emph{Weak Description}. A landmark description that is flawed, but not completely wrong (e.g. \emph{blue towel} described as \emph{blue napkin});
    \item \emph{Wrong Orientation}. An instruction refers to a landmark but orients it incorrectly with respect to the route, e.g. saying it is \textit{on the left side} when it is actually \textit{on the right side}.
    \end{compactitem}
\item \textbf{Path Errors}
    \begin{compactitem}
    \item \emph{Wrong action}. A mistake in an action/step (e.g. \emph{turn right} when one needs to \emph{turn left});
    \item \emph{Missing action}. Skipping a an action/step (e.g. the instruction neglects to mention a crucial right hand turn);
    \item \emph{Weak granularity}. Some segment of the instruction is too coarse to describe the multiple steps that are needed in the path trajectory (e.g. merely using \emph{go forward} to describe a route that passes through two doorways and a dining hall).
    \end{compactitem}
\end{compactitem}

\begin{figure*}[t]
\centering
\includegraphics[width=.9\linewidth]{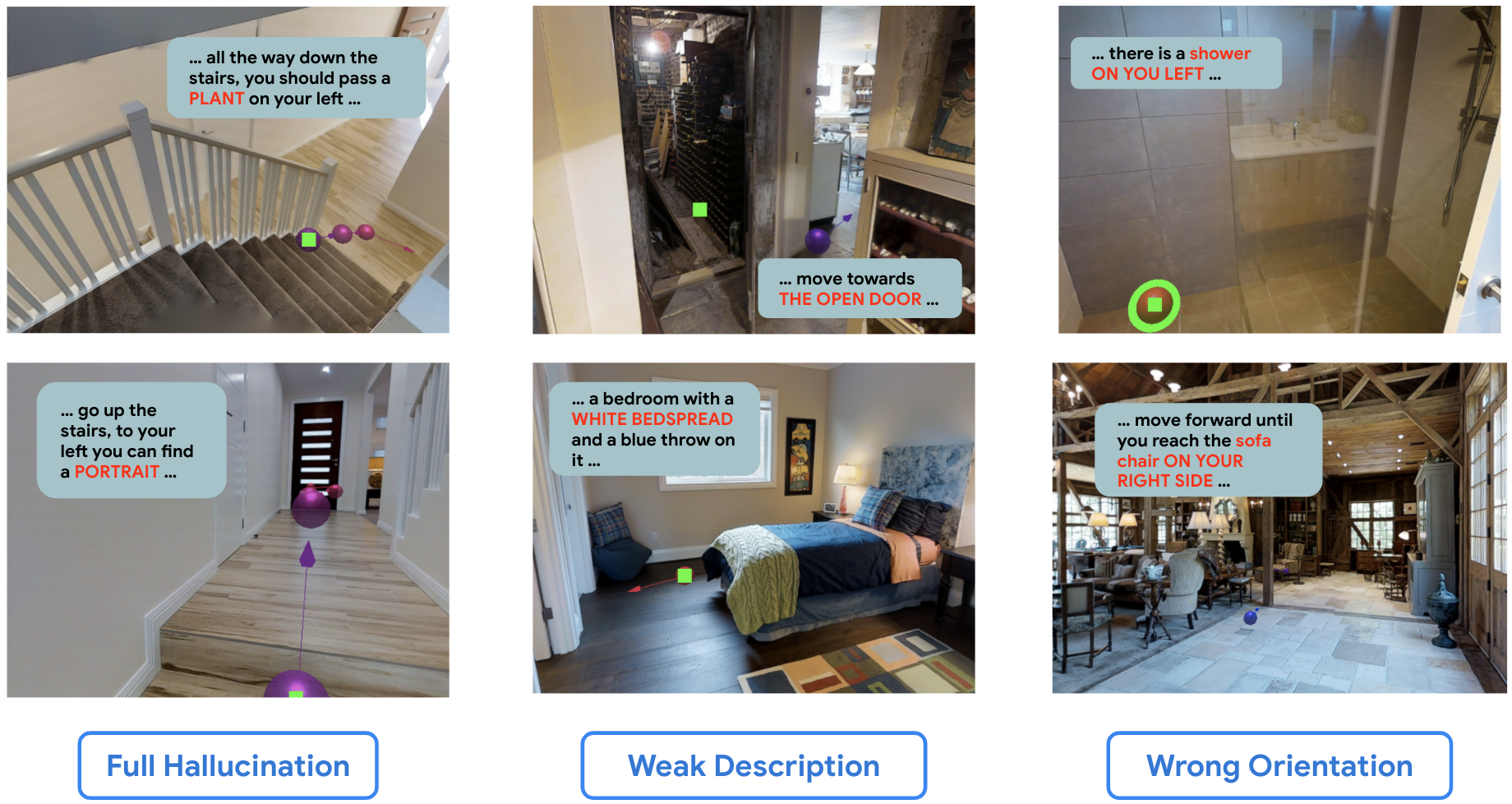}
\caption{Examples of \emph{Landmark Errors}. \textbf{Full Hallucination}: a \emph{plant} and \emph{portrait} are mentioned in the top and bottom panes respectively but these landmarks cannot be found in the visual scene. \textbf{Weak Description}: the top pane exemplifies an ambiguous landmark -- there are two open doors, and it's not clear which one to move towards; the bottom pane illustrates a flawed description (specifically, wrong color). \textbf{Wrong orientation}: On the top pane, the \emph{shower} is to the \emph{right hand side} of the wayfinder rather than left; in the bottom pane, the \emph{sofa chair} should be on the \emph{left side} instead.}
\label{fig:mp3d-landmark-errors}
\end{figure*}

\begin{figure*}[t]
\centering
\includegraphics[width=.9\linewidth]{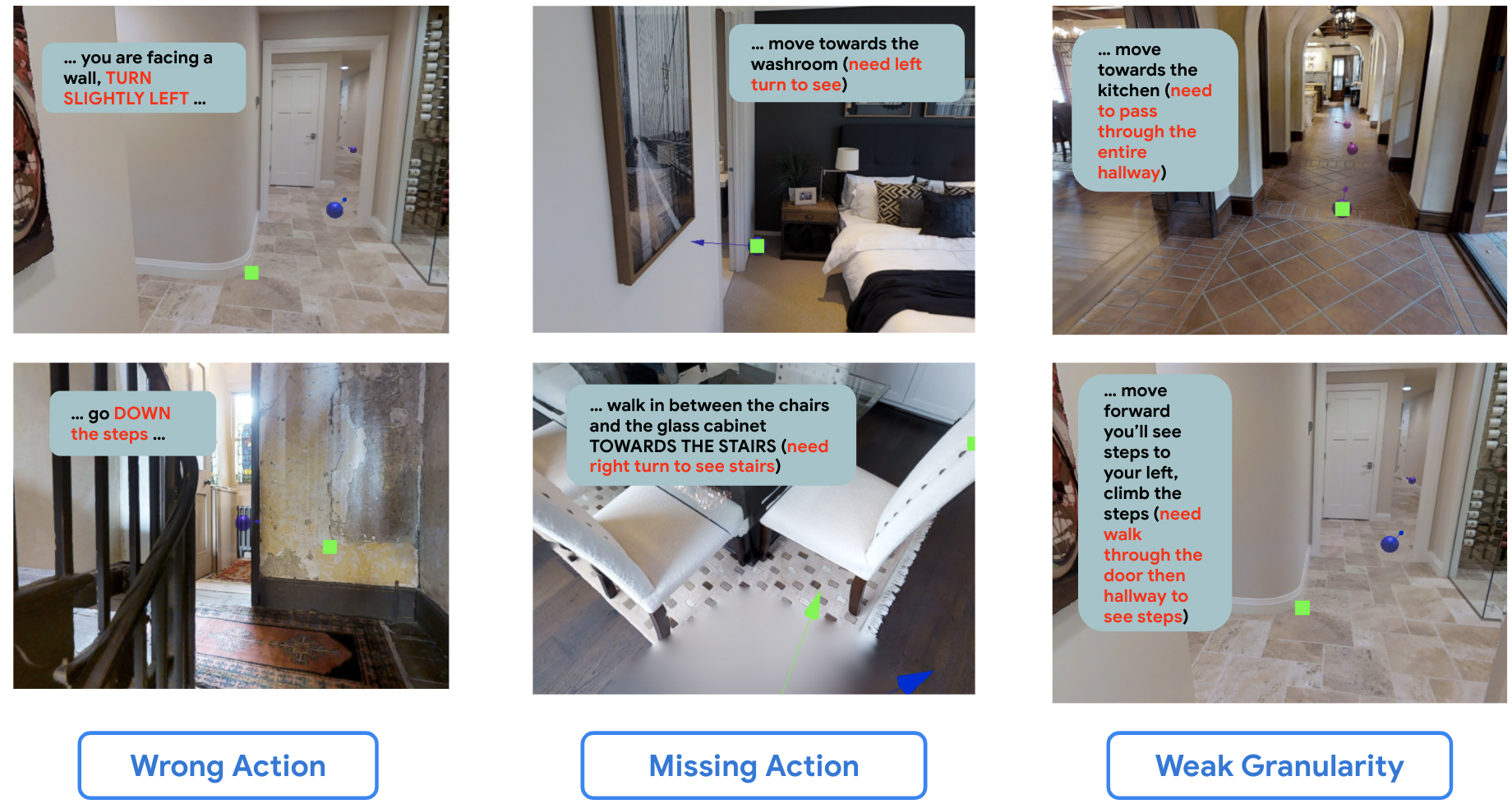}
\caption{Examples of \emph{Path Errors}. \textbf{Wrong action}: the examples show a right turn mistaken to be a left turn, and going up stairs mistaken as going down. \textbf{Missing action}: for the top pane, the \emph{washroom} is not visible before making an additional left turn; for the bottom pane, \emph{stairs} require a right turn to see. \textbf{Weak granularity}: in the examples, overly coarse instruction segments are given for long path segments.}
\label{fig:mp3d-path-errors}
\end{figure*}

We provide examples of each error type in \cref{fig:mp3d-landmark-errors} and \cref{fig:mp3d-path-errors}, where the blue/purple (arrowed) balls indicate the ground truth path. The error analysis was performed by the paper authors with a critical eye; any weakness in the instructions was annotated as an error.  

The results are summarized in \cref{tab:error-analysis-tally}.
Of the instructions annotated, 16\% were judged to be error-free, with some errors or weaknesses identified in the remaining 84\%. Out of the 6 error types, \textit{Weak Description} was by far the most common (62.9\% of all errors). In contrast, the proportion of \textit{Full Hallucination}, \textit{Wrong Orientation} and \textit{Wrong Action} errors--which were common in previous models--was relatively low, at 11.4\%, 7.9\% and 7.1\% respectively. 

Anecdotally, we noticed that the cost of different errors varies. Human wayfinders can often overcome minor flaws in the description of a landmark (e.g. if a \emph{\textbf{pink} bedspread} is misidentified as a \emph{\textbf{white} bedspread}).
However if ambiguity is involved, e.g. \emph{move towards \textbf{the open door}} when there more than one door is in view, confusion results. Full hallucination and wrong orientation can certainly be misleading but if the navigator survey the environment carefully in the context of the neighboring segments in the instruction, they are also often resolvable. The three types of path errors are less recoverable, as they often result in missteps that take the wayfinder to an entirely wrong path.

\begin{table}
    \centering
    \scalebox{.85}{
    \begin{tabular}{lr}
        & \% of all errors   \\
        \toprule
        \textbf{Landmark Errors:}\\
        Full Hallucination & 11.4 \\
        Weak Description & \textbf{62.9}  \\
        Wrong Orientation & 7.9 \\
        \textbf{Path Errors:}\\
        Wrong Action & 7.1 \\
        Missing Action & 3.6 \\
        Weak Granularity & 7.1 \\
    \end{tabular}}
    \caption{Of generated instructions with errors, 25.7\% of errors are issues with actions or landmark orientation (wrong, missing, or convoluted [i.e., weak granularity]).  Another 11.4\% of errors are full hallucinations but the overwhelming majority, 62.9\% are an issue with some aspect of the description of a landmark.}
    \label{tab:error-analysis-tally}
\end{table}

\paragraph{Results on automatic evaluation.} Beyond the strong quality of \ourmodel-generated instructions in human evaluation (\cref{tab:human-eval-r2r,tab:human-eval-rxr}), they also perform at the similar level in automatic/model evaluation. Employing the state-of-the-art HAMT~\cite{chen2021hamt} VLN agent (\cref{tab:vln-hamt-results}), in particular, we received 55.7\% vs. 56.5\% success rate and 63.3\% vs. 62.9\% nDTW between model-generated (with predicted landmarks) vs. human-written instructions, demonstrating that \ourmodel produces instructions followable by both human and model agents.

\begin{table}[!t]
    \centering
    \scalebox{0.80}{
    \small
    \setlength\tabcolsep{2.5pt}
    \begin{tabular}{llccccccc}
        & \textbf{Model} & \textbf{Landmarks} & \textbf{Aux} & \textbf{PT} & \textbf{SR} $\uparrow$ & \textbf{SPL} $\uparrow$ & \textbf{NDTW} $\uparrow$ & \textbf{SDTW} $\uparrow$   \\
        \toprule
        1 & SpkFol-RxR & Full Panos & & &29.6 & 25.9	& 41.6 & 23.4 \\
        2 & \ourmodel & Full Panos & & & 50.7	& 46.9	& 60.1 & 43.1 \\
        3 & \ourmodel & Outbound & & & 53.6 & 50.1 & 62.9 & 46.7 \\
        4 & \ourmodel & Silver & & & 55.9 & 52.1 & 64.1 & 48.6 \\
        5 & \ourmodel & Silver & \checkmark & & 56.3 & 52.3 & 64.2 & 48.9 \\
        6 & \ourmodel & Silver & \checkmark & \checkmark & 56.4 & 52.5 & 64.2 & 48.9 \\
        7 & \ourmodel & Pred. & \checkmark & \checkmark & 55.7 & 51.8 & 63.3 & 47.7 \\
        8 & Human  & & & & 56.5 & 52.7 & 62.9 & 48.4 \\
    \end{tabular}}
    \caption{Automatic evaluations of generated instructions on RxR Val-Unseen based on HAMT~\cite{chen2021hamt} wayfinding performance. Settings and row numbers correspond to Tab. 3 in the main paper.}
    \label{tab:vln-hamt-results}
\end{table}

\end{document}